\documentclass[acmsmall,screen]{acmart}

\usepackage{amsmath}
\usepackage{bbm}
\usepackage{booktabs} 
\usepackage{multirow}
\usepackage{subfigure}
\usepackage{hyperref}
\usepackage{breqn}
\usepackage{longtable}
\usepackage{mathtools}
\usepackage{algorithm}
\usepackage{algpseudocode}
\usepackage{paralist}
\usepackage{microtype}
\usepackage[detect-all]{siunitx}
\usepackage{cleveref}
\usepackage{lscape} 
\usepackage{upgreek}
\algrenewcommand\textproc{}
\usepackage{xcolor}
\newcommand{\method}{\texttt{FedSTAR}}
\DeclareMathOperator*{\argmax}{arg\,max}

\setcopyright{rightsretained}

\begin{document}

\acmJournal{TECS}
\acmYear{2022} \acmVolume{1} \acmNumber{1} \acmArticle{1} \acmMonth{1} \acmPrice{}\acmDOI{10.1145/3520128}

\title{Federated Self-Training for Semi-Supervised Audio Recognition}

\author{Vasileios Tsouvalas}
\affiliation{%
\institution{Eindhoven University of Technology}
\city{Eindhoven}
\country{The Netherlands}}
\email{v.tsouvalas@student.tue.nl}

\author{Aaqib Saeed}
\affiliation{%
\institution{Eindhoven University of Technology}
\city{Eindhoven}
\country{The Netherlands}}
\email{a.saeed@tue.nl}

\author{Tanir Ozcelebi}
\affiliation{%
\institution{Eindhoven University of Technology}
\city{Eindhoven}
\country{The Netherlands}}
\email{t.ozcelebi@tue.nl}

\renewcommand{\shortauthors}{Tsouvalas et al.}

\begin{abstract}
Federated Learning is a distributed machine learning paradigm dealing with decentralized and personal datasets. Since data reside on devices like smartphones and virtual assistants, labeling is entrusted to the clients or labels are extracted in an automated way. Specifically, in the case of audio data, acquiring semantic annotations can be prohibitively expensive and time-consuming. As a result, an abundance of audio data remains unlabeled and unexploited on users' devices. Most existing federated learning approaches focus on supervised learning without harnessing the unlabeled data. In this work, we study the problem of semi-supervised learning of audio models via self-training in conjunction with federated learning. We propose ~\method to exploit large-scale on-device unlabeled data to improve the generalization of audio recognition models. We further demonstrate that self-supervised pre-trained models can accelerate the training of on-device models, significantly improving convergence within fewer training rounds. We conduct experiments on diverse public audio classification datasets and investigate the performance of our models under varying percentages of labeled and unlabeled data. Notably, we show that with as little as 3\% labeled data available,~\method~on average can improve the recognition rate by 13.28\% compared to the fully-supervised federated model.
\end{abstract}


\maketitle
\section{Introduction}

The emergence of smartphones, wearables, and modern Internet of Things (IoT) devices results in a massive amount of highly informative data generated continuously from a multitude of embedded sensors and logs of user interactions with various applications. The ubiquity of these contemporary devices and the exponential growth of the data produced on edge provides a unique opportunity to tackle critical problems in various domains, such as healthcare, well-being, manufacturing, and infrastructure monitoring. Notably, the advent of deep learning has enabled us to leverage these raw data directly for learning models while leaving ad-hoc (hand-designed) approaches largely redundant. The improved schemes for learning deep networks and the availability of massive labeled datasets have brought tremendous advancements in several areas, including language modeling, audio understanding, object recognition, image synthesis, and more.

Traditionally, developing machine learning models or performing analytics in a data center context requires the data from IoT devices to be pooled or aggregated in a centralized repository before processing it further for the desired objective. However, the rapidly increasing size of available data, in combination with the high communication costs and possible bandwidth limitations, render the accumulation of data in a cloud-based server unfeasible~\citep{FederatedSurvey}. Additionally, such centralized data aggregation schemes could also be restricted by privacy issues and regulations (e.g., General Data Protection Regulation). Due to these factors and the growing computational and storage capabilities of distributed devices, it is appealing to leave the data decentralized and perform operations directly on the device that collects that data through primarily utilizing local resources.

The rapidly evolving Federated Learning (FL) field is concerned with distributed training of machine learning models on the decentralized data residing on remote devices like smartphones and wearables. The key idea behind FL is to bring the computation (or code) closer to where the data reside to harness data locality extensively. Specifically, in a federated setting, minimal updates to the models (e.g., parameters of a neural network) are performed entirely on-device and communicated to the central server, which aggregates these updates from all participating devices to produce a unified global model. Unlike the standard way of learning models, the salient differentiating factor is that the data never leaves the user's device, which is an appealing property for privacy-sensitive data. This strategy has been applied on a wide range of tasks in recent years \citep{KeywordSpottingIID, KeywordSuggestion, KeyboardPrediction, FederatedEmojiPrediction}. Nevertheless, a common limitation of existing approaches is that they primarily focus on a supervised learning regime. The implicit assumption that the labeled data is widely available on the device, or it can be easily labeled through user interaction or programmatically, such as for keyword prediction or photo categorization, is in most pragmatic cases unrealistic.

\begin{figure}[t]
    \centering \small
    \includegraphics[clip,  width=0.9\textwidth]{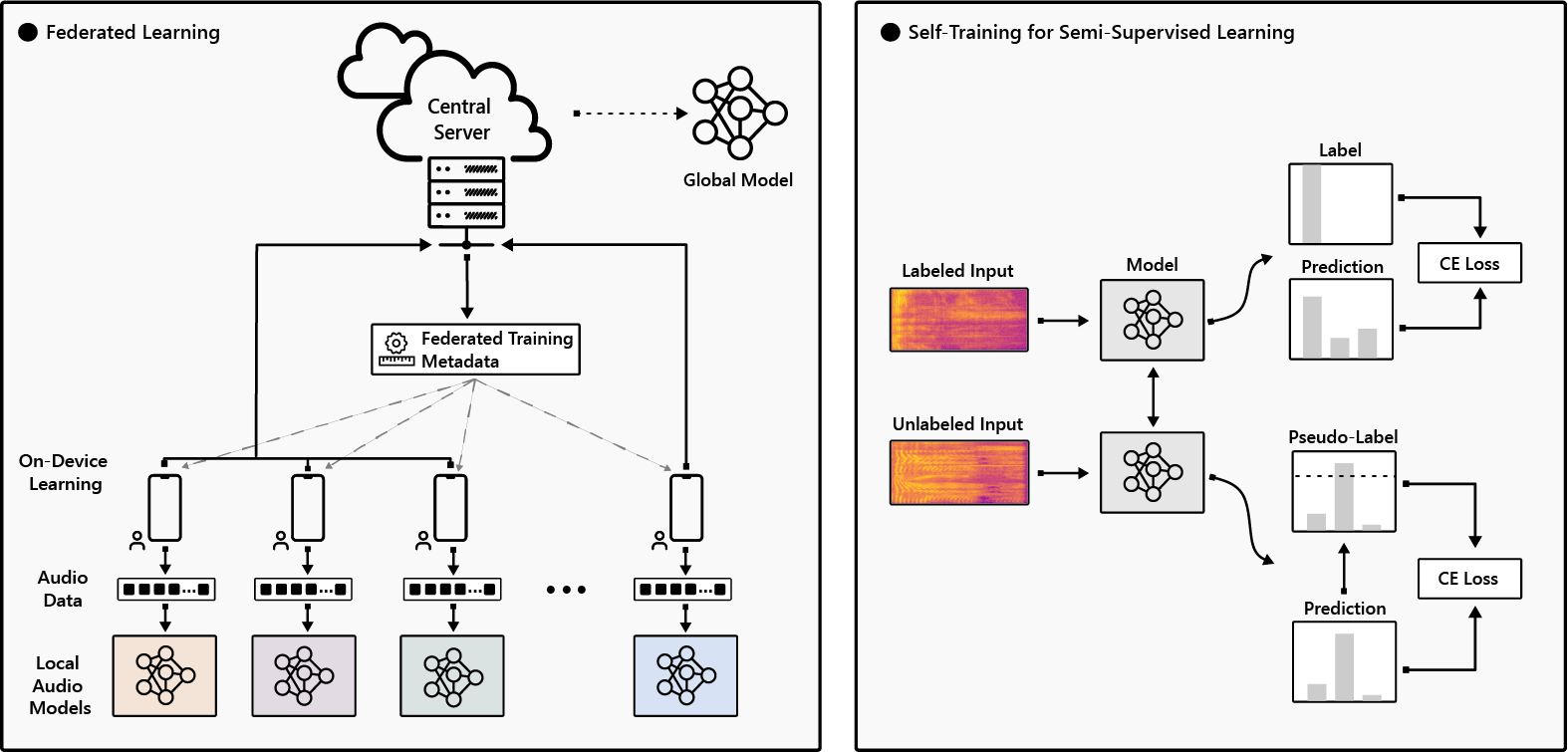}
    \caption{Illustration of~\method~ for label-efficient learning of audio recognition models in a federated setting.} \label{fig:overview}
\end{figure}

In reality, on-device data is largely unlabeled and constantly expanding in size. It cannot be labeled to the same extent as standard datasets, which are annotated via crowd-sourcing or other means for training deep neural networks. Due to the prohibitive cost of annotation, users have little to no incentives, and notably for various important tasks, the domain knowledge missing to perform the annotation process appropriately leaves most of the data residing on devices to remain unlabeled. This is especially true when considering the utilization of audio data to perform various audio recognition tasks, which have recently attracted increasing interest from researchers. As a result, numerous audio recognition systems have been developed, such as for wildlife monitoring~\citep{BirdDetection,BatDetection} and surveillance~\citep{RoadSurveillance}. In addition to monitoring applications, highly accurate acoustic models are utilized for keyword spotting for virtual assistants~\citep{KeywordSpottingIID}, anomaly detection for machine sounds~\citep{MachineAnomalyDetection}, and in the development of health risk diagnosis systems, such as cardiac arrest detection~\citep{chan2019contactless}. However, in the majority of such applications, there is no straightforward manner for the annotation process. For instance, suppose that we have a sleep tracker application that assesses a person's risk of obstructive sleep apnea using breathing and snoring sounds during sleep. In this case, the end-users may not be able to evaluate their sleeping sounds sufficiently, and clinicians may need to analyze and annotate the samples. Even in cases where no human expertise is required, like in a music tagging application, the correct labeling of songs requires effort on the user's end. Additionally, there are cases where distributed devices host models with no human-in-the-loop to annotate the audio data, such as surveillance devices, making the labeling process infeasible. Thus, in many realistic scenarios for FL, local audio data will be primarily unlabeled. This leads to a novel FL problem, namely \textit{semi-supervised federated learning}, where users' devices collectively hold a massive amount of unlabeled audio samples and only a fraction of labeled audio examples. \par

Semi-supervised learning techniques have been widely deployed in a centralized learning setting to utilize readily available unlabeled data, and could also be applied in federated learning settings. In particular, with semi-supervision of models, available unlabeled data can be exploited during the training phase, improving the overall performance of the resulting model~\citep{SSLSurvey}. Pseudo-labeling is a widely applied semi-supervised learning method, which relies on the predictions of a model on unlabeled data, i.e., pseudo labels, to utilize unlabeled data during the learning phase~\citep{PSL}. With no structural requirements from the input modalities and tiny computational overhead, pseudo-labeling is an ideal candidate to be applied in federated learning settings, where device heterogeneity and computational resources vary across devices. To this end, we propose a federated self-training approach, named ~\method~ (Federated Self-Training for Audio Recognition), to unify semi-supervision with federated learning to leverage large-scale unlabeled audio data. With the exploitation of unannotated audio samples that reside on clients' devices, we aim to improve the generalization of federated models on a wide range of audio recognition tasks under a pragmatic scenario, where scarcity of labels poses a significant challenge for learning useful models.

Apart from the labels' deficiency, FL introduces other challenges of the system and statistical heterogeneity~\citep{OpenChallenges}. These challenges lead to device hardware and data collection diverseness that can significantly affect the number of devices participating in each federated round as well as the on-device data distribution. Several FL techniques provide flexibility in selecting a fraction of clients in each training round and address the non-i.i.d. nature of client's data distributions, such as FedAvg~\citep{FedAvg} and FedProx~\citep{FedProx}. The training convergence properties of such distributed optimization methods are discussed in~\citep{OpenChallenges}, where a clear reduction in the convergence rates is reported. In a centralized setting, self-supervised pre-training can improve the model's convergence and generalization through leveraging pre-training on massive unlabeled datasets~\cite{saeed2021contrastive}. With self-supervised learning, the model is able to learn useful representations from unlabeled data; thus, when used for the downstream task, self-supervised model can significantly improve the training efficiency and predictive performance~\cite{saeed2021contrastive}. To address the issue of slow training convergence in federated settings, we propose the utilization of self-supervised pre-trained models as model initialization for the FL procedure as compared to the naive random initialization of model parameters. Through extensive evaluation, we demonstrate that the convergence rate of our proposed semi-supervised federated algorithm, i.e.,~\method, can be greatly improved by using a pre-trained model learned in a self-supervised manner. \par

To the best of our knowledge \method~is the first FL approach that learns models for audio recognition tasks by utilizing not only labeled but also unlabeled samples on user devices while not being dependent on any data (labeled or unlabeled) on the server side. Just like the labeled samples, the on-device unlabeled samples are utilized locally by self-training based on our proposed pseudo-labeling with dynamic prediction confidence thresholding. As \method~is not altering either the utilized model's architecture or the global model averaging process, the underlined hardware requirements are similar to the chosen federated learning algorithm (e.g., FedAvg), while the on-device storage demand is unaffected, since~\method~essentially uses already stored unlabeled data that are left unexploited. In addition, with the utilization of unlabeled data, \method~models are less sensitive to the non-i.i.d. nature of the labeled data across clients (label distribution skew and data sample imbalance across clients). As a result, it performs much better in typical non-i.i.d. data federated settings. Furthermore, solutions in the literature focus on randomly initialized models at the server side. We for the first time employs self-supervised pre-training on the server side using a publicly available audio data to further improve the efficiency of training, which means fewer training rounds are needed for convergence.

Concisely, the main contributions of this work are as follows:
\begin{itemize}
    \item We study on the practical problem of semi-supervised federated learning for audio recognition tasks to address the lack of labeled data that presents a major challenge for learning on-device models.

    \item We design a simple yet effective approach based on self-training, called ~\method. It exploits large-scale unlabeled distributed data in a federated setting with the help of a novel adaptive confidence thresholding mechanism for effectively generating pseudo-labels.

    \item We exploit self-supervised models pre-trained on FSD-50K corpus~\citep{fonseca2020fsd50k} for significantly improving training convergence in federated settings.

    \item We demonstrate through extensive evaluation that our technique is able to effectively learn generalizable audio models under a variety of federated settings and label availability on diverse public datasets, namely Speech Commands~\citep{SPCM}, Ambient Context~\citep{ABCX} and VoxForge~\citep{VF}.

    \item We show that ~\method, with as few as 3\% labeled data, on average can improve recognition rate by 13.28\% across all datasets compared to the fully-supervised federated models.

\end{itemize}

The rest of the paper is organized as follows. In Section 2, an overview of the related work is provided. Section 3 presents an overview of related paradigms and methodologies as background information, Section 4 introduces the proposed federated self-training approach for semi-supervised audio recognition. Section 5 presents an evaluation of \method~on publicly available datasets. Finally, Section 6 concludes the paper and lists future directions for research.

\section{Related Work}
\textbf{Federated Learning.} FL has been attracting growing attention thanks to its unique characteristic of collaboratively training machine learning models without actually sharing local data and compromising users' privacy~\citep{FL}. The most popular and simplistic approach to learning models from decentralized data is the Federated Averaging (FedAvg) algorithm~\citep{FedAvg}. Specifically, FedAvg performs several local stochastic gradient descent (SGD) steps on a sampled subset of devices' data in parallel and aggregates the locally learned model parameters on a central server to generate a unified global model through weighted averaging. This strategy has proved to work relatively well for a wide range of tasks in i.i.d. settings~\citep{KeywordSpottingIID,KeywordSuggestion}. At the same time, the performance can decrease substantially when FedAvg is exposed to non-i.i.d. data distribution~\citep{NonIID,OpenChallenges}. Authors in ~\citep{NonIID} proposed globally sharing a portion of the dataset to improve FL performance under non-i.i.d settings. In addition to the challenge introduced by data distribution, communication efficiency is another critical problem in FL. The communication challenges could be alleviated by increasing the number of local SGD steps between sequential communication stages. However, with the increase of SGD steps, the device's model may begin to diverge, and the aggregation of such models can affect the generalization of global models~\citep{FedProx}. FedProx was proposed to tackle this issue by adding a loss term to restrict the local models' updates to be closer to the existing global model~\citep{FedProx}. Nevertheless, a typical limitation of existing work is the focus on a supervised learning regime with the implicit assumption that the local private data is fully labeled or could be labeled simplistically through labeling functions. However, in the majority of pragmatic scenarios, a straightforward annotation process is non-existent.  \par 

Recently, performing on-device federated training of acoustic models has attracted considerable attention~\citep{KeywordSpottingIID,KeywordSuggestion,KeywordSpottingNonIID,TrainEESR,TrainUA}. In \citep{KeywordSpottingIID}, FL was employed for a keyword spotting task and the development of a wake-word detection system, whereas,~\citep{TrainEESR,KeywordSpottingNonIID} investigated the effect of non-i.i.d. distributions on the same task. In~\citep{TrainEESR}, a highly skewed data distribution scenario was considered, where a large set of speakers used their devices to record a set of sentences. To address the challenges introduced due to the non-i.i.d. distribution of data, a word-error-rate model aggregation strategy was developed. In addition, a training scheme with a centralized model, pre-trained on a small portion of the dataset, was also examined. Furthermore,~\citep{KeywordSpottingNonIID} considered a scenario where devices might hold unlabeled audio samples and used a semi-supervised federated scheme based on a teacher-student architecture to exploit unlabeled audio data. However, the teacher model relied on additional high-quality labeled data for training in a centralized setting. Likewise, \citep{TrainUA} introduced a framework for privacy-preserving training of user authentication models with FL using labeled audio data. Nonetheless, all prior approaches consider only semantically annotated audio examples or require supplementary labeled data on the server-side to utilize the available unlabeled audio data that reside on devices. To address these problems, we propose a self-training approach to exploit unlabeled audio samples residing on clients' devices. In addition, as servers often possess the computational resources to efficiently pre-train a model on a massive unlabeled dataset, we employ self-supervision to develop a model that can be used as a highly-effective starting point for federated training instead of using randomly initialized weights. \par

\textbf{Semi-Supervised Learning.} In semi-supervised learning (SSL), we are provided with a dataset containing both labeled and unlabeled examples, where the labeled fraction is generally tiny compared to the unlabeled one and the curation of strong labels for the unlabeled dataset is impractical due to time constraints, cost, and privacy-related issues~\citep{SSL}. While there is a wide range of SSL methods and approaches that have been developed in the area of deep learning, we will mainly focus on the self-training or pseudo-labeling approach~\citep{PSL}. Self-training uses the prediction on unlabeled data to supervise the model's training in combination with a small percentage of labeled data. Specifically, pseudo-labels are constructed by extracting one-hot labels from highly confident predictions on unlabeled data. These are then used as training targets in a supervised learning regime. This simplistic approach of utilizing unlabeled data has been combined with various methods to further improve the training efficiency. In~\cite{DeepPSL}, authors demonstrated that setting a minimum number of labeled samples per training batch can be effective to reduce over-fitting due to noise accumulation on generated predictions. In addition, the use of a scalar temperature for scaling softmax output achieves a softer probability distribution over classes for the predictions and urges models to generate the correct pseudo-labels without suffering from over-confidence~\cite{DistillingPSL}. This temperature scaling approach can be highly beneficial in modern deep neural networks architectures, which have shown to suffer from over-confident predictions~\cite{DistillingPSL}. Supplementary, MixMatch proposed sharpening the prediction's distribution to further improve the generated pseudo-labels predictions~\cite{MixMatch}. The sharpening process is performed by averaging the predictions' distribution of augmentation versions of the same unlabeled sample. Apart from self-training, alternative SSL approaches introduce a loss term, which is computed on unlabeled data, to encourages the model to generalize better to unseen data. Based on the objective of the loss term, we can classify these approaches in two categories: consistency regularization techniques - which are based on the principle that a classifier should produce the same class distribution for an unlabeled sample even after augmentation~\cite{VAT,MT}; and entropy minimization techniques - which aim to motivate the model to produce low-entropy (high-confident) predictions for all unlabeled data~\cite{Entropy}. For a concise review and realistic evaluation of various deep learning based semi-supervised techniques, we refer an interested reader to~\cite{SSLRealisticEvaluation}.\par

A recent study~\citep{SSFL} has questioned the soundness of the assumption that devices have well-annotated labels in a federated setting. Existing semi-supervised federated learning (SSFL) approaches, such as FedMatch~\citep{FedMatch} and FedSemi~\citep{FedSemi}, have only recently started to be examined under the vision domain to exploit unlabeled data. FedMatch decomposes the parameters learned from labeled and unlabeled on-device data and uses an inter-client consistency loss to enforce consistency between the pseudo-labeling predictions made across multiple devices. In \citep{FedSemi}, FedSemi adapts a mean teacher approach to harvest the unlabeled data and proposes an adaptive layer selection to reduce the communication cost during the training process. Apart these methods, many studies consider different data distribution schemes, including sharing an unlabeled dataset across devices~\citep{ShareUnlabelledDataset}. Lastly, it is important to note that recent works employ SSFL to address problems in healthcare domain, namely, electronic health records \citep{FederatedMedical} and for problems like human activity recognition \citep{ActivityRecognition}. Nevertheless, none of the discussed approaches focuses on learning models for audio recognition tasks by utilizing devices' unlabeled audio samples. \par

\section{Background}

In this section, we provide a brief overview of semi-supervised and federated learning paradigms as they act as fundamental building blocks of our federated self-training approach for utilizing large-scale on-device unlabeled audio data in a federated setting. 

\subsection{Semi-Supervised Learning} \label{ssection:ssl}
Given enough computational power and supervised data, deep neural networks have proven to achieve human-level performance on a wide variety of problems~\citep{DeepLearningPerformance}. However, the curation of large-scale datasets is very costly and time-consuming as it either requires crowd-sourcing or domain expertise, such as in the case of medical imaging. Likewise, for several practical problems, it is simply not possible to create a large enough labeled dataset (e.g., due to privacy issues) to learn a model of reasonable accuracy. In such cases, SSL algorithms offer a compelling alternative to fully supervised methods for jointly learning from the fraction of labeled and a large number of unlabeled instances.  \par

Specifically, SSL aims to solve the problem of learning with partially labeled data where the ratio of unlabeled training examples is usually much larger than that of the labeled ones. Formally, let $\mathcal{D}_{L} = \left \{ \left ( x_{l_i},y_{i} \right ) \right \}_{i=1}^{N_{l}}$ represent a set of labeled data, where $N_{l}$ is the number of labeled data, $x_{l_i}$ is an input instance,  $y_{i} ~\epsilon \left \{ 1, \cdots, \mathcal{C} \right \}$ is the corresponding label, and $\mathcal{C}$ is the number of label categories for the  $\mathcal{C}$-way multi-class classification problem. Besides, we have a set of unlabeled samples denoted as $\mathcal{D}_{U} = \left \{ x_{u_i}  \right \}_{i=1}^{N_{u}}$, where, $N_{u}$ is the number of unlabeled data. Let $p_\theta\left(y \mid x \right)$ be a neural network that is parameterized by weights $\theta$ that predicts softmax outputs $\widehat{y}$ for a given input $x$. In the setting of semi-supervised learning, where in general $N_{l} \ll N_{u}$, we need to simultaneously minimize losses on both labeled and unlabeled data to learn the model's parameters $\theta$. Specifically, our objective is to minimize the following loss function:

\begin{equation} \label{eqn:SSL}
    \mathcal{L}_{\theta} = \mathcal{L}_{s_{\theta}}(\mathcal{D}_{L}) +  \mathcal{L}_{u_{\theta}}(\mathcal{D}_{U})
\end{equation}

\noindent where $\mathcal{L}_{s_{\theta}}(\mathcal{D}_{L})$ and $\mathcal{L}_{u_{\theta}}(\mathcal{D}_{U})$ are the loss terms from supervised and unsupervised learning, respectively. 

The teacher-student self-training framework is a popular scheme to simultaneously learn from both labeled and unlabeled data. In this approach, we firstly use the available labeled data to train a good teacher model, which is then utilized to label any available unlabeled data. Consequently, both labeled and unlabeled data are used to jointly train a student model. In this way, the model assumes a dual role as a teacher and a student. In particular, as a student, it learns from the available data, while as a teacher, it generates targets to help the learning process of student. Since the model itself generates targets, they may very well be incorrect, thus, the learning experience of the student model depends solely on the ability of teacher model to generate high-quality targets~\citep{teacher_Student}.

\subsection{Federated Learning}
FL is a novel collaborative learning paradigm that aims to learn a single, global model from data stored on remote clients with no need to share their data with a central server. In particular, with the data residing on clients' devices, a subset of clients is selected to perform a number of local SGD steps on their data in parallel in each communication round. Upon completion, clients exchange their models' weights updates with the server, aiming to learn a unified global model by aggregating these updates. Formally, the goal of FL is typically to minimize the following objective function:

\begin{equation} \label{eqn:FL}
    \min_{\theta} \mathcal{L}_{\theta} = \sum_{k=1}^{K} \gamma_{k} {\mathcal{L}}_k(\theta)
\end{equation}

\noindent where $\mathcal{L}_k$ is the minimization function of the $k$-th client and $\gamma_{k}$ corresponds to the relative impact of the $k$-th client to the construction of the global model. For the FedAvg algorithm, parameter $\gamma_{k}$ is equal to the ratio of client's local data $N_k$ over all training samples $\left (\gamma_{k} = \frac{N_k}{N}\right)$. \par

Specifically, let $\mathcal{D} = \left \{ \left ( x_{l_i},y_{i} \right ) \right \}_{i=1}^{N}$ be a dataset of $N$ labeled examples, similarly to the previously discussed dataset $\mathcal{D}_{L}$ in Section~\ref{ssection:ssl}. Given $K$ clients, $\mathcal{D}$ is decomposed into $K$ sub-datasets $\mathcal{D}^{k}=\left \{ \left (x_{l_i},y_{i} \right) \right \}_{i=1}^{N_k}$ corresponding to each clients' privately held data. For an initial global model $G$, the $\mathit{r}$-th communication round starts with server randomly selecting a portion $q$ ($0<q\leq K$) of clients to participate in the current training round. Afterwards, each client's local model receives the global parameters $\theta_r^G$ and performs supervised learning on their local dataset $\mathcal{D}^{k}$ to minimize $\mathcal{L}_{k}( \theta_r^k)$. Subsequently, $G$ aggregates over locally updated parameters by performing $\theta_{r+1}^G  \leftarrow   \sum\nolimits_{i=1}^{q} \frac{N_i}{N} \theta_r^i$. The presented circular training process, comprising of model weights' exchanges between server and clients, repeats until $\theta^G$ converges after $R$ rounds. \par

\section{Methodology}

In this section, we present our federated self-training learning approach, namely \method, for audio recognition tasks. First, we provide a formal overview of the problem, which \method~ aims to solve. Next, we discuss the proposed self-training technique (i.e., pseudo-labeling with dynamic prediction confidence thresholding) in detail, followed by the presentation of our \method~ algorithm. Finally, we provide a thorough description of the self-supervised pre-training technique used to train a model as an initialization point for the \method~ approach. 

\subsection{Problem Formulation} \label{problem_definition}
We focus on the problem of \textit{SSFL}, where labeled data are scarce across users' devices. At the same time, clients collectively hold a massive amount of unlabeled audio data. 
In addition, in a typical federated learning setting, the on-device data distribution depends on the profile of the users operating the devices. Thus, it is a common scenario for both labeled and unlabeled data to originate from the same data distribution. Based on the aforementioned assumption, with~\method, we aim to utilize the available unlabeled data on clients and further improve the generalization of FL models, alleviating the need for clients to hold well-annotated data. In this way, we substantially decouple the amount of available labeled from the predictive power of acoustic models trained under federated settings. \par

Formally, under the setting of \textit{SSFL}, each of the $K$ clients holds a labeled set, $\mathcal{D}_{L}^{k} = \{ \left ( x_{l_i},y_{i} \right ) \}_{i=1}^{N_{l,k}}$ and an unlabeled set $\mathcal{D}_{U}^{k} = \{x_{u_i}\}_{i=1}^{N_{u,k}}$, where $N_{k} = N_{l,k} + N_{u,k}$ is the total number of data samples stored on the $\mathit{k}$-th client and $N_{l,k} \ll N_{u,k}$. We desire to learn a global unified model $G$ without clients sharing any of their local data, $\mathcal{D}_L^k$ and $\mathcal{D}_U^k$. To this end, our objective is to simultaneously minimize both supervised and unsupervised learning losses during each client's local training step on the $\mathit{r}$-th round of the FL algorithm. Specifically, the minimization function, similar to the one presented in Equation~\ref{eqn:FL}, is:

\begin{equation} \label{eqn:SSFL}
    \min_{\theta} {\mathcal{L}}_{\theta} = \sum_{k=1}^{K} \gamma_{k} {\mathcal{L}}_k(\theta) \textrm{ where } \mathcal{L}_{k}(\theta) =\mathcal{L}_{s_{\theta}}(\mathcal{D}_{L}^{k}) + \beta \mathcal{L}_{u_{\theta}}(\mathcal{D}_{U}^{k})
\end{equation}

\noindent Here $\mathcal{L}_{s}(\mathcal{D}_{L}^{k})$ is the loss terms from supervised learning on the labeled data held by the $k$-th client, and $\mathcal{L}_{u}(\mathcal{D}_{U}^{k})$ represents the loss term from unsupervised learning on the unlabeled data of the same client. We add the parameter $\beta$ to control the effect of unlabeled data on the training procedure, while $\gamma_{k}$ is the relative impact of the $k$-th client on the construction of the global model $G$.\par

\begin{figure}[h]
    \centering \small
    \includegraphics[width=1.0\textwidth]{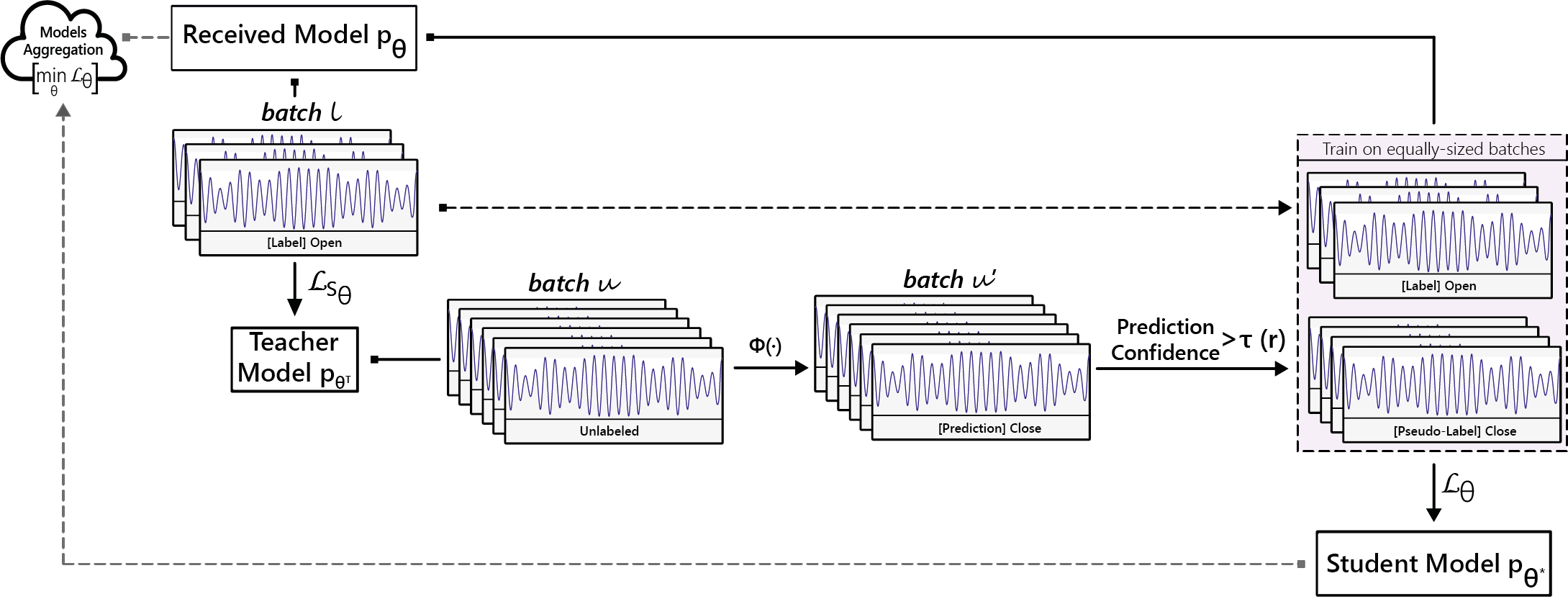}
    \caption{On-device self-training based on pseudo-labeling in a federated setting for an audio recognition task  shown for illustration proposes.} \label{fig:overview_sec_4}
\end{figure}

\subsection{Self-Training with Pseudo Labeling}
Self-training via pseudo-labeling has been widely used in semi-supervised learning~\citep{SSLSurvey}. The objective of highly effective teacher-student self-training approaches is to train a teacher model, which supervises the learning process of a student model that learns from labeled and unlabeled data jointly. Firstly, a teacher model is built with the available labeled data and afterwards this is exploited to make predictions for the unlabeled samples. Subsequently, the student model is trained on both labeled and predicted samples. We propose a self-training technique with a dynamic prediction confidence threshold to learn from the unlabeled audio data residing on the client's device, thus boosting the performance of models trained in federated settings with varying percentages of labeled examples. For audio classification tasks, in order to learn from the labeled datasets $\mathcal{D}_{L}^k$ across all participating clients, we apply cross-entropy loss as follows:

\begin{equation} \label{eqn:SCE}
    \begin{aligned}
        \mathcal{L}_{s}(\mathcal{D}_{L}^{k}) = - \frac{1}{N_{l,k}} \sum\limits_{i=1}^{N_{l,k}}\sum\limits_{j=1}^{C} y_{i}^{j} \log (\mathit{f}_{i}^{\theta^{k}}(x_{l_j})) = \mathcal{L}_{CE}\left ( y,p_{\theta^k}\left(y\mid x_l \right) \right )
    \end{aligned}
\end{equation}

Next, to learn from unlabeled data, we generate pseudo-labels $\widehat{y}$ for all available unlabeled data $x_{u}$ on client $\mathit{k}$ by performing:

\begin{equation} \label{eqn:PSL}
    \begin{aligned}
        \widehat{y} = \Upphi \left ( z,T \right ) =   \argmax\limits_{i \in  \left \{ 1, \dots, \mathcal{C} \right \}} \left (\frac{e^{z_i/T}}{\sum_{j=1}^{\mathcal{C}} e^{z_j/T}} \right )
    \end{aligned}  
\end{equation}

\noindent where $z_{i}$ are the logits produced for the input sample $x_{u_i}$ by the $k$-th client model $p_{\theta^k}$ before the softmax layer. In essence, $\Upphi$ produces categorical labels for the given “\textit{soften}” softmax values, in which temperature scaling is applied with a constant scalar temperature $T$. As the maximum of the softmax function remains unaltered, the predicted pseudo-label $\widehat{y}$ is identical as if the original prediction (without scaling) for an unlabeled sample $x_u$ was used; however, the prediction confidence is weakened. A dynamic threshold $\tau$ of confidence is proposed following a cosine schedule to discard low-confidence predictions when generating pseudo-labels. For the obtained pseudo-labels, we then perform standard cross-entropy minimization while using $\widehat{y}$ as targets in the following manner:   

\begin{equation} \label{eqn:UCE}
    \begin{aligned}
        \mathcal{L}_{u}(\mathcal{D}_{U}^{k}) &= - \frac{1}{N_{u,k}} \sum\limits_{i=1}^{N_{u,k}}\sum\limits_{j=1}^{C} \widehat{y_{i}}^{j} \log (\mathit{f}_{i}^{\theta^{k}}(x_{u_j})) = \mathcal{L}_{CE}\left ( \widehat{y},p_{\theta^k}\left(x_u\right) \right )
    \end{aligned}
\end{equation}

Revising the initial minimization goal of \method~ expressed in Equation~\ref{eqn:SSFL}, we can now represent local models' loss function on the $\mathit{r}$-th round of the FL algorithm for the $\mathit{k}$-th client as:

\begin{equation} \label{eqn:SSFL_detail}
    \begin{aligned}
        \mathcal{L}_{k}(\theta^k) = \mathcal{L}_{CE} \left ( y,p_{\theta^k}\left(y\mid x_l\right) \right ) + \beta \mathcal{L}_{CE} \left ( \widehat{y},p_{\theta^k}\left( x_u\right) \right)
    \end{aligned}
\end{equation}

\subsection{Federated Self-training} \label{sec:self_supervised_learning}
The objective of federated self-training is to create a teacher model on each client to exploit labeled data resident on clients' devices, which will be used to predict labels for the unlabeled instances available in the device. As both labeled and unlabeled on-device samples originate from the same data distribution, a student model can be constructed on each client device by collectively training on labeled and pseudo-labeled data, the weights of whom will be returned to the server for aggregation. Under federated settings, however, a more complicated analysis is required, as clients' local labeled data can be limited and can have a highly skewed distribution. In such settings, teacher models may produce inaccurate pseudo-label predictions, and student classifiers potentially amplify the mistakes further during training through using faulty pseudo-labels. To ensure the proper construction of pseudo-labels and guarantee that the student model will learn properly from unlabeled data, the confidence of the predictions is taken into consideration when generating pseudo-labels to discard any low-confidence predictions. \par

Concisely, in the proposed~\method~algorithm, the clients' local update step is altered to learn from unlabeled datasets $\mathcal{D}_{U}^{k}$. As can be seen in Figure~\ref{fig:overview_sec_4}, a representative round $\mathit{r}$ of~\method~starts with the distribution of global models' weights $\theta^G$ to a randomly selected subset of $q$ clients. On each client, equally sized batches $l$ and $u$ from the labeled and unlabeled sets are created, respectively. The model's weights update is performed, as in Equation~\ref{eqn:SSFL_detail}. Firstly, the classical supervised categorical cross-entropy loss is minimized for batch $l$, as in Equation~\ref{eqn:SCE} to construct a teacher model, and afterwards, with this model pseudo-labels are produced using $\Phi(\cdot)$. With the creation of pseudo-labels $\widehat{y}$, the unlabeled batch $u$ is then treated as a labeled batch $u^{'} =\left \{ \left ( u,\widehat{y} \right ) \right \}$, in which the client's model is further trained with standard cross-entropy minimization. It is important to note that we simultaneously optimize the cross-entropy loss on both $l$ and $u$ subsets by computing both losses before performing backpropagation to update the local models' parameters. Lastly, the locally updated weights from all participating clients in the $\mathit{r}$-th round are sent back to the server, where the global model weights are calculated as a weighted average over all the local weights updates.\par

Since $N_{l} \ll N_{u}$ holds for all clients, given a sufficient number of participating rounds, unlabeled instances will be exposed to all the available labeled data. Additionally, we propose an \textit{adaptive confidence thresholding} method to diminish unsatisfactory performance due to training on faulty pseudo-labels. In particular, in addition to using temperature scaling $T$ to ``\textit{soften}'' softmax output and generated confident predictions, we employ an increasing confidence threshold $\tau$ to discard low-confidence pseudo-labels during training following a cosine schedule. Cosine learning rate schedulers rely on the observation that we might not want to decrease the learning rate too drastically in the beginning, while we might want to “refine” our solution in the end using a very small learning rate. Along the same lines, with our cosine confidence thresholding, we allow clients to explore the locally-stored unlabeled data, $D_{U}^{k}$, in the first few federated rounds, while considering only highly-confident predictions in a later stage of the training procedure. While other methods could be explored for this purpose, such a study is outside the scope of the current work and we mainly focus on cosine scheduler, which has proven to work well empirically across a variety of tasks~\citep{cosine_sceduler}. Further details and an overview of our approach for the semi-supervised training procedure can be found in Algorithm~\ref{alg:PSL}.\par

\begin{algorithm}[t]
\caption{\method: Federated Self-training for Audio Recognition. In the algorithm, $l$ and $u$ are equally sized batches from on-device labeled and unlabeled samples respectively. Scalar $\beta$ controls the affect of unlabeled data in the training process, and $\eta$ is the learning rate.} \label{alg:PSL}
\begin{algorithmic}[1]
    \State Server initialization of model $G$ with model weights $\theta_{0}^{G}$
    \For{ $i=1,\dots,R$ }
        \State Randomly select $K$ clients to participate in round $i$
        \For{ each client $k \in K$ \textbf{ in parallel}}
            \State $\theta_{i}^{k} \gets \theta_{i}^{G}$
            \State $\theta_{i+1}^{k} \gets$ ClientUpdate($\theta_{i}^{k}$) 
        \EndFor
        \State $\theta_{i+1}^{G} \gets \sum\nolimits_{k=1}^{K} \frac{N_k}{N} \theta_{i+1}^k$
    \EndFor
    \Procedure{ClientUpdate}{$\theta$}
        \For{epoch $e=1,2,\dots,E$}
            \For{ batch $l \in \mathcal{D}_{L}$ and $u \in \mathcal{D}_{U}$}
                \State $\widehat{y} \gets \Upphi \left( p_{\theta}(x_u),T \right)$
                \State $\theta \gets \theta - \eta \nabla_{\theta} \left(  \mathcal{L}_{CE}(y,p_{\theta}(y \mid x_l)) + \beta \cdot \mathcal{L}_{CE}(\widehat{y},p_{\theta}(x_u))\right)$ 
            \EndFor
        \EndFor
    \EndProcedure
\end{algorithmic}
\end{algorithm}

\subsection{Self-Supervised Pretraining Strategy} 
\label{sec:pre_train}
Self-supervised learning aims to learn useful representations from unlabeled data by tasking a model to solve an auxiliary task for which supervision can be acquired from the input itself. Given an unlabeled data $D = \{x\}_{m=1}^M$ and deep neural network $f_{\theta}(.)$, the aim is to pre-train a model through solving a surrogate task, where, labels $y$ for the standard objective function (e.g., cross-entropy) are extracted automatically from $x$. The learned model is then utilized as a fixed feature extractor or as initialization for rapidly learning downstream tasks of interest. The fields of computer vision and natural language processing have seen tremendous progress in representation learning with deep networks in a self-learning manner, with no human intervention in the labeling process. Here, the prominent techniques for audio representation learning from unlabeled data include and audio-visual synchronization~\cite{korbar2018cooperative}, contrastive learning~\cite{saeed2021contrastive}, and other auxiliary tasks~\cite{Model}.\par

In our work, we propose to leverage self-supervised pre-training on the server side to improve training convergence of~\method~on client devices. Motivated by the fact that the server can often hold a large amount of unlabeled data and has enormous computational resources available, we employ contrastive learning for audio to develop a model that can be used as an effective starting point for federated learning instead of using randomly initialized weights. Specifically, pre-training is performed in a centralized setting with a separate publicly available dataset on the server side; thus it can be done once and be used repeatedly for different downstream tasks. To the best of our knowledge, this is the first time self-supervised learning has been used to address the convergence of federated models with a fewer training rounds efficiently. \par 

Formally, we pre-train our model with contrastive learning~\citep{saeed2021contrastive} using FSD-$50$K~\citep{fonseca2020fsd50k} dataset. On a high level, the objective is to train a model to maximize the similarity between related audio segments while minimizing it for the rest. Similar samples are generated through stochastic sampling from the same audio clip, while other segments in a batch are treated as negatives. In particular, we use bilinear similarity formulation and pre-train our model with a batch size of $1024$ for $500$ epochs. Moreover, we utilize a network architecture, as described in Section~\ref{sec:model_architecture} as an encoder with the addition of a dense layer containing $256$ hidden units on top, which is discarded after the pre-training stage. In this way by using a same architecture, we are able to draw proper conclusions for the effects of utilizing a pre-trained model as an initial global model and directly compare with the randomly initialized~\method~approach.

\section{Experiments}
In this section, we conduct an extensive evaluation of our approach on publicly available datasets for various audio recognition tasks to determine the efficacy of \method~ in learning generalizable models under a variety of federated settings and label availability. Firstly, the federated learning framework and datasets utilized for validation are presented, followed by a detailed description of the network architecture. Next, we introduce our experiments in centralized and fully supervised federated settings, which serve as a baseline for evaluating our approach. Finally, we provide a thorough evaluation of \method~, which is structured in the form of several research questions.

\subsection{Datasets and Audio Pre-Processing}
We use publicly available datasets to evaluate our models on a range of audio recognition tasks. For all datasets, we use the suggested train/test split for comparability purposes. For ambient sound classification, we use the Ambient Acoustic Contexts dataset~\citep{ABCX}, in which sounds from ten distinct events are present. For the keyword spotting task, we use the second version of the Speech Commands dataset~\citep{SPCM}, where the objective is to detect when a particular keyword is spoken out of a set of twelve target classes. Likewise, we use VoxForge~\citep{VF} for the task of spoken language classification, which contains audio recordings in six languages - English, Spanish, French, German, Russian, and Italian. It is one of the largest available datasets for language identification problems; it is valuable for benchmarking the performance of the supervised FL model. We resampled the Ambient Acoustic Contexts samples from 48 kHz to 16 kHz to utilize the same sampling frequency across all our datasets samples. In Table ~\ref{tab:datasets}, we present a description of each dataset.\par

\begin{table}[!htbp]
\centering \small
\caption{Key details of the datasets used in evaluation.} \label{tab:datasets}
\begin{tabular}{@{}lcc@{}}
    \toprule
    \multicolumn{1}{c}{\textbf{Dataset}} & \textbf{Task} & \textbf{Classes} \\ 
    \midrule[0.5pt] 
    Ambient Context~\citep{ABCX}        & Event classification      & 10 \\
    Speech Commands~\citep{SPCM}        & Keyword spotting          & 12 \\
    VoxForge~\citep{VF}                 & Language identification   & 6 \\ 
    \bottomrule
\end{tabular}%
\end{table} 

\subsection{Model Architecture and Optimization} \label{sec:model_architecture}
The network architecture of our global model is inspired by \citep{Model} with a key distinction that instead of batch normalization, we utilize group normalization \citep{GNorm} after each convolutional layer and employ a spatial dropout layer. We use log-Mel spectrograms as the model's input, which we compute by applying a short-time Fourier transform on the one-second audio segment with a window size of $25$ \textit{ms} and a hop size equal to $10$ \textit{ms} to extract $64$ Mel-spaced frequency bins for each window. In order to make an accurate prediction on an audio clip, we average over the predictions of non-overlapping segments of an entire audio clip. Our convolutional neural network architecture consists of four blocks. In each block, we perform two separate convolutions, one on the temporal and another on the frequency dimension, outputs of which we concatenate afterward in order to perform a joint $1 \times 1$ convolution. Using this scheme, the model can capture fine-grained features from each dimension and discover high-level features from their shared output. Furthermore, we apply L$2$ regularization with a rate of $0.0001$ in each convolution layer and group normalization \citep{GNorm} after each layer. Between blocks, we utilize max-pooling to reduce the time-frequency dimensions by a factor of two and use a spatial dropout rate of $0.1$ to avoid over-fitting. We apply ReLU as a non-linear activation function and use Adam optimizer with the default learning rate of $0.001$ to minimize categorical cross-entropy.\par

To simulate a federated environment, we use the Flower framework~\citep{Flower} and utilize FedAvg~\citep{FedAvg} as an optimization algorithm to construct the global model from clients' local updates. Additionally, a number of parameters were selected to control the federated settings of our self-training strategy fully. Those parameters are: \begin{inparaenum}[1)] 
\item $N$ - number of clients, 
\item $R$ - number of rounds, 
\item $q$ - clients' participation percentage in each round, 
\item $E$ - number of local train steps per round, 
\item $\sigma$ - data distribution variance across clients,
\item $L$ - dataset's percentage to be used as labeled samples,
\item $U$ - dataset's percentage to be used as unlabeled samples (excluding L\% of the data used as labeled),
\item $\beta$ - influence of unlabeled data over training process, 
\item $T$ - temperature scaling parameter, and
\item $\tau$ - predictions confidence threshold.
\end{inparaenum}
We employ uniform random sampling for the clients' selection strategy, as other approaches for adequate clients election are outside the current work scope. Lastly, across all \method~ experiments, we fixed the temperature scaling parameter $T=4$, while we set the confidence threshold $\tau$ to initialize from $0.5$ and gradually increase to a maximum of $0.9$ during training, following a cosine schedule. A description of the parameters used is presented in Table ~\ref{tab:paramters}. \par

\begin{table}[!htbp]
\centering \small
\caption{Primary Experiment Parameters.} \label{tab:paramters}
\begin{tabular}{@{}lcc@{}}
    \toprule
    \multicolumn{1}{c}{\textbf{Parameter Name}} & \textbf{Variable} & \textbf{Range} \\ 
    \midrule[0.5pt] 
    Number of Clients                               & $N$       & 5 -- 30       \\
    Number of Federated Rounds                      & $R$       & 1 -- 100      \\
    Number of Local Train Steps                     & $E$       & 1 -- 4        \\
    Clients' Participation Percentage               & $q$       & 20\% to 80\%  \\
    Data Distribution Variance across Clients       & $\sigma$  & 0\% to 50\%   \\
    Dataset's Labeled Percentage                    & $L$       & 3\% to 100\%  \\
    Dataset's Utilized Unlabeled Percentage         & $U$       & 20\% to 100\% \\
    Unlabeled Data Influence on Train Step          & $\beta$   & 50\%          \\
    Temperature Scaling                             & $T$       & 4             \\
    Confidence Threshold Percentage                 & $\tau$    & 50\% to 90\%  \\
    \bottomrule
\end{tabular}%

\end{table} 

\subsection{Baselines and Evaluation Strategy} \label{sec:baseline}
In fully supervised federated experiments where the complete dataset is available, the labeled instances are randomly distributed across the available clients. Likewise, in experiments where the creation of a labeled subset from the original dataset is required ($L<$100\%), we keep the dataset's initial class distribution ratio to avoid tempering with dataset characteristics. Afterward, the labeled subset is again randomly distributed across the available clients. With the $\sigma$ parameter set to 25\% and a random partitioning of labeled samples among clients, the labeled data distribution resembles a non-i.i.d. one. In contrast, an increase of available clients results in a highly skewed distribution. It is worth mentioning that even if the meaning of non-i.i.d. is generally straightforward, data can be non-i.i.d in many ways. In our work, the term non-i.i.d data distribution describes a distribution with both a label distribution skew and a quantity skew (data samples imbalance across clients). This type of data distribution is common across clients' data in federated settings. Each client frequently corresponds to a particular user (affecting the label distribution), and the application usage across clients can differ substantially (affecting the label distribution). For a concise taxonomy of non-i.i.d. data regimes, we refer our readers to~\citep{OpenChallenges}. Additionally, in \method~, the unlabeled subset consists of the dataset's remaining samples after extracting the provided labels. In such experiments, both the labeled and unlabeled subsets are dispensed at random over the available clients. Furthermore, for an accurate comparison between our experiments, we manage any randomness during the data partitioning and training procedures by passing a seed alongside the parameters presented in Table~\ref{tab:paramters}. In this way, we control the amount of data and the data instances that reside on each simulated client. Lastly, for a more rigorous evaluation, we perform three distinct trials (or runs, i.e., training a model from scratch) in each setting, and the average accuracy over all three runs is reported across the results of Sections \ref{sec:baseline} and \ref{sec:results}. \par

To evaluate the \method, we first need to construct a high-quality \textit{supervised} baseline both in centralized and federated environments. Therefore, we perform preliminary experiments in both centralized as well as fully-supervised federated settings. We conduct initial experiments on all datasets in centralized settings where the models are trained until convergence to obtain the resulting accuracy on a test set, which is presented in the centralized row of Table~\ref{tab:baseline}. Following, we examine our model's performance in federated settings by adjusting the FL parameters to $R$=100, $q$=80\%, $E$=1 and $\sigma$=25\%. We vary the number of clients ($N$) while keeping the remaining federated parameters to the same as the earlier mentioned values as $N$ frequently fluctuates in real-life FL scenarios. Thus, we can explore how the federated model behaves as clients progressively increase and the available local data become yet more distributed, affecting the performance of FL~\citep{NonIID}. The results for supervised FL are presented in the Federated row of Table~\ref{tab:baseline}. We note from results presented in Table~\ref{tab:baseline} that the supervised federated models achieve comparable results in various cases to the models trained in a centralized setting across all three datasets. Moreover, the number of clients ($N$) has a clear effect on the model's performance. With an increase in $N$, we notice that the training process requires more training rounds ($R$) to converge as the quantity of local data for each client decreases. The obtained accuracy for a constant number of rounds deteriorates. \par

\subsection{Results} \label{sec:results}
\subsubsection{\textbf{Comparison against fully-supervised federated approach under non-i.i.d. settings.}} \label{sec:comparison} 

We first evaluate \method~ to determine the obtained improvements versus a fully-supervised federated approach when a non-i.i.d. distribution is considered. This analysis helps in understanding \textit{whether utilizing unlabeled instances that reside on clients' devices with \method~ can be beneficial for a model trained in federated settings and, if so, to which degree it improves the recognition rate}. To this end, we perform experiments on all three datasets for a diverse number of clients ($N$) where the percentage of available labeled instances is varied from $3\%$ up to $50\%$. To clearly illustrate the performance gain of \method~ in comparison to the supervised FL regime, experiments with identically labeled subsets are conducted under fully-supervised FL, where, the unlabeled instances remained unexploited. Table~\ref{tab:results} provides the accuracy scores on test sets averaged across three independent runs for the considered datasets to be robust against differences in weight initialization and optimization. For ease of comparison, we add the results column on fully-supervised FL using entire labeled dataset ($L$=100\%) in Table~\ref{tab:results}, as discussed earlier in Section~\ref{sec:baseline}.\par

\begin{table}[t]
\centering \small
\caption{Evaluation of audio recognition models in centralized and fully-supervised federated settings. Average accuracy on test set over three distinct trials. Federated parameters are set to $q$=80\%, $\sigma$=25\%, $L$=100\%, $E$=1, $R$=100.} \label{tab:baseline}
\begin{tabular}{lp{0.5cm}ccc} 
    \toprule
    \multicolumn{2}{c}{\textbf{Method}} & \begin{tabular}[c]{@{}c@{}}\textbf{Speech} \textbf{Commands}\end{tabular} & \begin{tabular}[c]{@{}c@{}}\textbf{Ambient} \textbf{Context}\end{tabular} & \textbf{VoxForge} \\ 
    \midrule[0.5pt] 
    \multicolumn{1}{l}{Centralized} & & 96.54 & 73.03 & 79.60 \\
    \cmidrule[0.2pt](rl){1-5}
    \multicolumn{1}{l}{\multirow{4}{*}{Federated}}
        & $N$=5  & 96.93 & 71.88 & 79.13 \\
        & $N$=10 & 96.78 & 68.01 & 78.98 \\
        & $N$=15 & 96.33 & 66.86 & 76.09 \\
        & $N$=30 & 94.62 & 65.14 & 65.17 \\
    \bottomrule
\end{tabular}%
\end{table}
 
In Table~\ref{tab:results}, we observe that \method~ can utilize unlabeled audio data to improve the model's performance across all datasets significantly. Consequently, we can conclude that \method~ can be applied in a federated environment with scarce labeled audio instances to boost the performance by learning from unlabeled data, independent of the audio recognition task. In particular, comparing the two rows for L=3\%, we note an increase of 13.28\% in accuracy on average when using \method~across the considered tasks.

While varying \textit{L}, we note that the percentage gab between \method~and the supervised federated counterparts shrinks as more labeled data are available across devices. In addition, with only $5$\% of labels available, we note that \method~model's accuracy is within a reasonable range from the ones trained under fully-supervised federated settings, where the complete dataset is available ($L$=100\%). These two observations suggest that~\method~can be especially useful under extreme label scarcity scenarios, where a highly accurate model can be obtained though the exploitation of unlabeled data. Alternatively, \method~could also be used in cases where sufficient labels are provided ($L$=50\%) to slightly improve the resulting models' performance. An exception to the aforementioned behavior is the case of \method~ experiments on VoxForge with $L$=50\%, where we can see that the accuracy obtained after $R$=100 rounds is inferior to the one achieved with identical federated settings and $L$=20\%. This might be because the unlabeled subset was not yet exposed to all available labeled examples; thus, the model had not reached the learning plateau and may require more training rounds to converge. Despite this behavior, the performance of \method~ models is superior to the supervised federated models with the same amount of labeled data and reasonably close to supervised federated models trained on the entire labeled data.

While $N$ increases and the labeled subset of each client shrinks (and hence we obtain an even higher non-i.i.d. distribution), we notice that the \method~ models' accuracy remains relatively unaffected, especially if we recollect that in FL experiments of Table~\ref{tab:baseline} we noticed accuracy decays for a constant $R$ as $N$ rises. In particular, when $N$=30 and the data distribution becomes highly skewed (or non-i.i.d), we note a performance gap between \method~ and supervised FL that can reach up to 50\% for the Speech Commands dataset. It means that the \method~ model can effectively utilize unlabeled audio data, even in highly distributed scenarios. This essentially means, that the exploitation of large-scale unlabeled on-device instances can help create a more uniform distribution across devices and tackle the challenges introduced in federated settings from the “non-i.i.d.-ness” of data.

\begin{table*}[t]
\centering \small
\caption{Performance evaluation of \method. Average accuracy over 3 distinct trials on test set. Detailed results are given in Table~\ref{tab:results_std} of the Appendix. Federated parameters are set to $q$=80\%, $\sigma$=25\%, $\beta$=0.5, $E$=1, $R$=100.} \label{tab:results}
\resizebox{\textwidth}{!}{%
    \begin{tabular}{l p{1cm} ccccccccc}
        \toprule
    
        \multicolumn{1}{c}{\multirow{2}{*}{\textbf{Dataset}}} 
        & \multicolumn{1}{c}{\multirow{2}{*}{\textbf{Clients}}}
        & \multicolumn{5}{c}{\textbf{Supervised (Federated)}} 
        & \multicolumn{4}{c}{\textbf{FedSTAR}} \\ 
        \cmidrule[0.5pt](rl){3-7} \cmidrule[0.5pt](rl){8-11}
    
        & & $L=3\%$ & $L=5\%$ & $L=20\%$ & $L=50\%$ & $L=100\%$ 
        & $L=3\%$ & $L=5\%$ & $L=20\%$ & $L=50\%$ \\ 
    
        \midrule[0.5pt]
        Ambient Context & \multicolumn{1}{c}{\multirow{3}{*}{$5$}}
                            & 46.34 & 47.89 & 61.40 & 65.85 & 71.88
                            & \textbf{48.68} & \textbf{54.95} & \textbf{64.37} & \textbf{67.04} \\
    
        Speech Commands &   & 81.12 & 87.97 & 92.35 & 94.66 & 96.93
                            & \textbf{87.41} & \textbf{90.01} & \textbf{94.17} & \textbf{94.85} \\
    
        VoxForge        &   & 54.55 & 56.41 & 61.65 & \textbf{70.37} & 79.13
                            & \textbf{63.92} & \textbf{67.80} & \textbf{69.09} & 67.08 \\
        \midrule[0.5pt]
        Ambient Context & \multicolumn{1}{c}{\multirow{3}{*}{$10$}}
                            & 35.29 & 41.31 & 51.71 & 62.69 & 68.01
                            & \textbf{48.87} & \textbf{52.37} & \textbf{62.94} & \textbf{64.42} \\
                            
        Speech Commands &   & 67.75 & 83.80 & 92.12 & 94.02 & 96.78
                            & \textbf{86.82} & \textbf{90.33} & \textbf{94.09} & \textbf{94.18} \\
                            
        VoxForge        &   & 56.14 & 54.73 & 60.48 & 62.41 & 78.98
                            & \textbf{59.87} & \textbf{64.35} & \textbf{69.38} & \textbf{63.27} \\
        \midrule[0.5pt]
        Ambient Context & \multicolumn{1}{c}{\multirow{3}{*}{$15$}}
                            & 33.03 & 42.75 & 53.37 & 59.97 & 66.86
                            & \textbf{49.54} & \textbf{54.71} & \textbf{63.46} & \textbf{62.41} \\
                            
        Speech Commands &   & 62.98 & 72.84 & 92.14 & 93.14 & 96.33
                            & \textbf{86.82} & \textbf{89.33} & \textbf{93.16} & \textbf{93.39} \\
                            
        VoxForge        &   & 54.26 & 54.37 & 57.11 & 60.29 & 76.09
                            & \textbf{55.82} & \textbf{57.96} & \textbf{67.66} & \textbf{61.66} \\
        \midrule[0.5pt]
        Ambient Context & \multicolumn{1}{c}{\multirow{3}{*}{$30$}}
                            & 32.31 & 40.17 & 47.05 & 55.85 & 65.14
                            & \textbf{40.84} & \textbf{46.58} & \textbf{60.21} & \textbf{56.19} \\
                            
        Speech Commands &   & 33.78 & 44.21 & 84.94 & 92.21 & 94.62
                            & \textbf{83.88} & \textbf{88.19} & \textbf{92.92} & \textbf{92.62} \\
                            
        VoxForge        &   & 50.32 & 54.33 & 55.19 & 57.56 & 65.17
                            & \textbf{54.81} & \textbf{56.18} & \textbf{63.83} & \textbf{56.66} \\
        \bottomrule
    \end{tabular}%
}
\end{table*}

\subsubsection{\textbf{Effectiveness of \method~ across diverse federated settings.}} \label{sec:robustness} In this subsection, we assess the efficacy of \method~ across a variety of federated settings. As presented in Section~\ref{sec:comparison}, the performance improvement by utilizing \method~ in comparison to the supervised FL scheme can vary across different federated settings. As federated settings' variability is a primary characteristic of a distributed environment, it is essential for our approach to be effective in distinct scenarios. To this end, we conduct further experiments on the Speech Command dataset, where the participation rate ($q$), the number of clients ($N$), the local train step ($E$), and the data distribution across clients ($\sigma$) are varied. We choose to investigate these four parameters as they primarily vary in a real-life FL setting, and they can have a significant effect on model performance~\citep{NonIID,OpenChallenges}. 

\begin{figure}[h]
    \centering \small
    \includegraphics[width=0.75\textwidth]{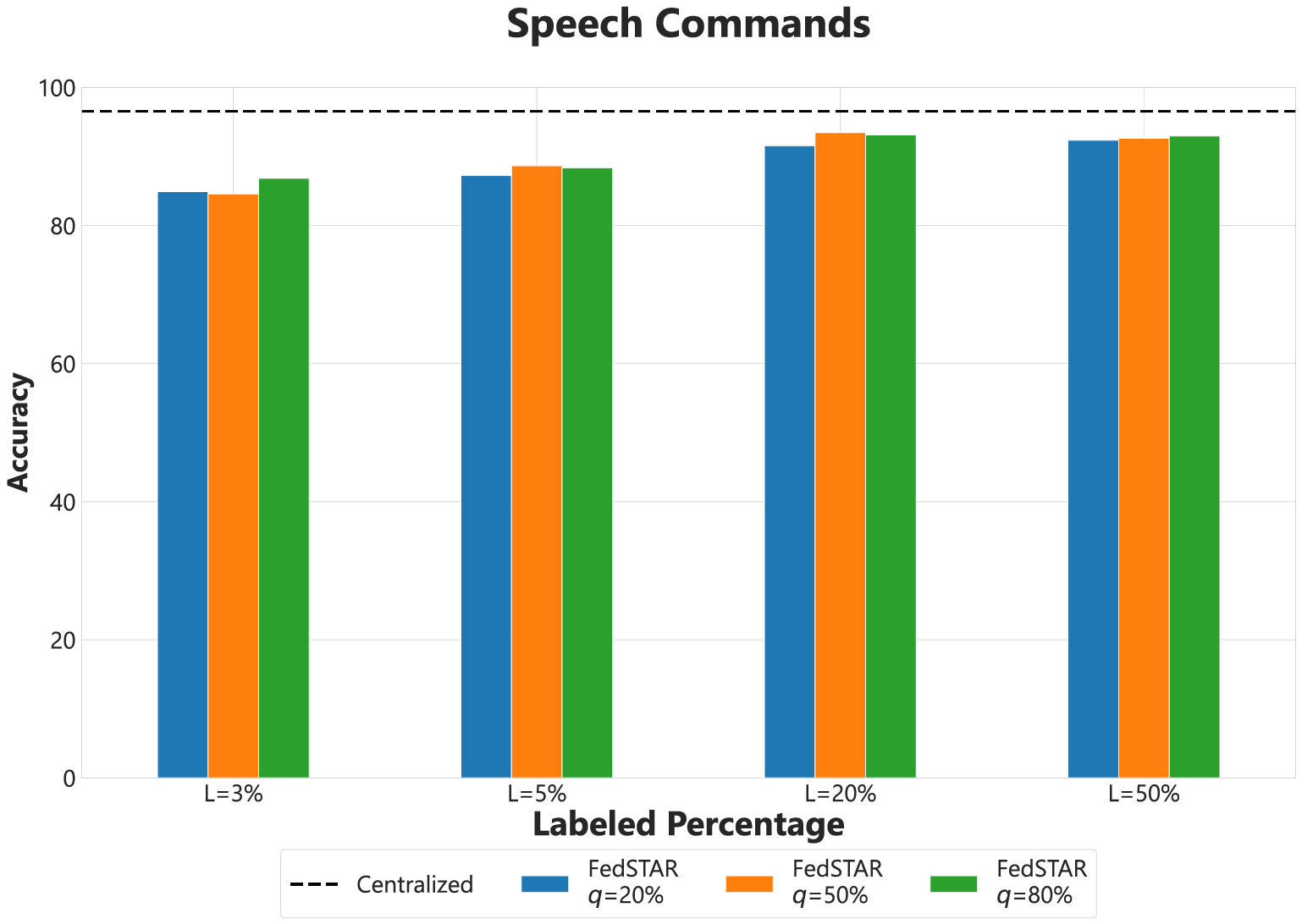}
    \caption{Evaluation of \method~ performance under varying clients' participation rate. Federated parameters are set to $\sigma$=25\%, $\beta=0.5$, $R$=100, $E$=1 and $N$=15. Average accuracy over three distinct trials is reported.} \label{fig:robustness_q}
\end{figure}

\textbf{Varying participation rate:} With the device heterogeneity and computational resources significantly varying across devices in a federated environment, a participation rate of 100\% is probably an unrealistic assumption for most pragmatic FL applications~\citep{OpenChallenges}. As clients' participation rate ($q$) can greatly influence the convergence rate of an FL model, we evaluate \method~ performance while varying the participation rate in each federated round. Therefore, this assessment helps us in \textit{understanding whether \method~ can retain the same level of effectiveness under low levels of clients participation.} To this end, we conduct experiments with $N$=15 for various clients' participation ($q$) rates, starting from 20\% up to 80\%, under different percentages of labels availability on the Speech Commands dataset. The Figure~\ref{fig:robustness_q} provide obtained accuracy score on the test set; we observe that \method~ is able to effectively learn from the unlabeled instances residing on clients' devices under low levels of clients engagement, even if the available labeled samples are scarce. While there is a decrease in \method~ model's accuracy when the participation rate reduces, the reduction is no more than 2\% for a given $L$. In particular, the reduction is eliminated when additional labeled instances are available. \par

\begin{figure}[h]
    \centering \small
    \includegraphics[width=0.75\textwidth]{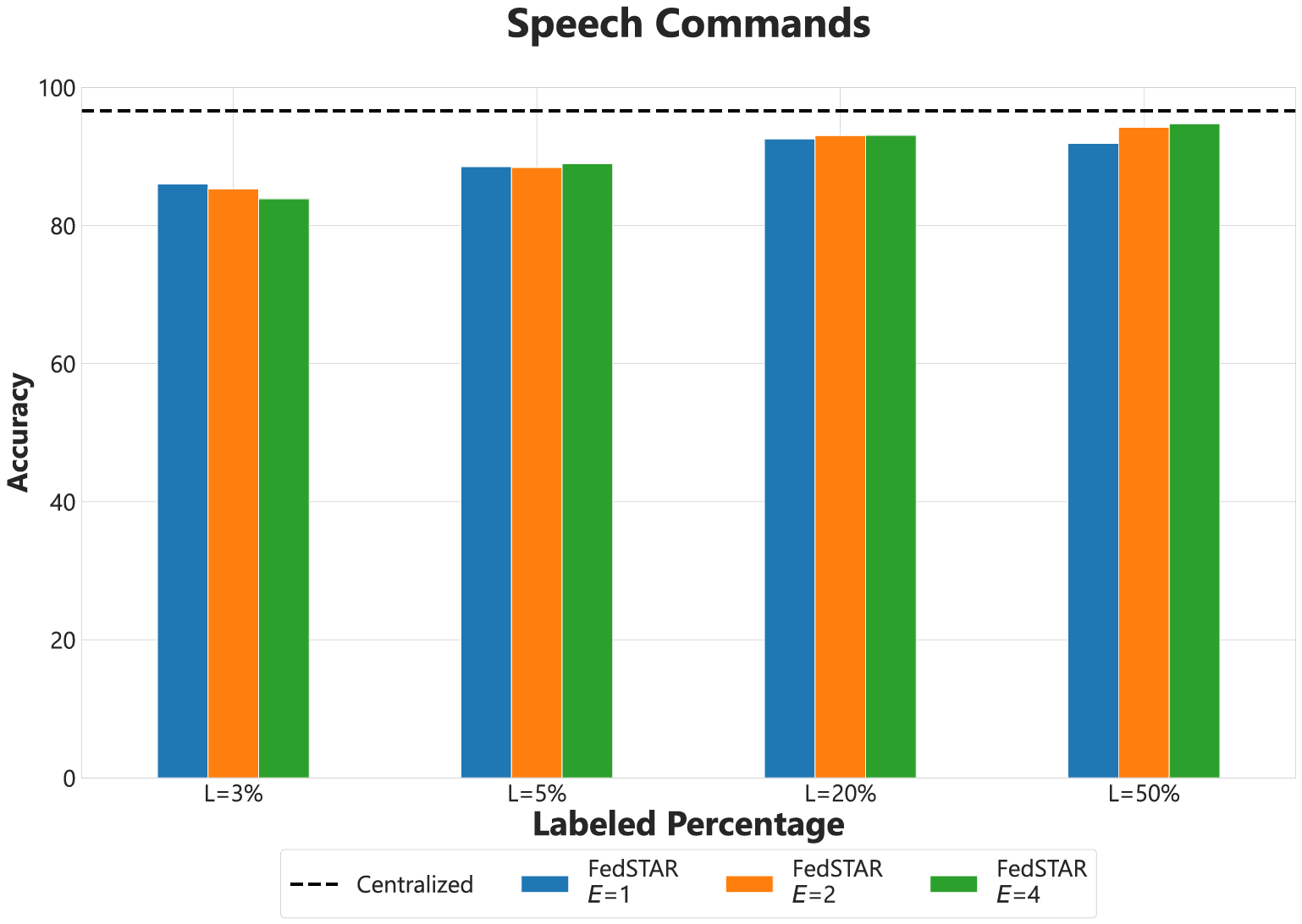}
    \caption{Evaluation of \method~performance against local train steps size. Federated parameters are set to $\sigma$=25\%, $\beta=0.5$, $R$=50, $q$=80\% and $N$=15. Average accuracy over 3 distinct trials is reported.} \label{fig:robustness_E}
\end{figure}

\textbf{Varying local train steps:} Subsequently, we examine the effect of increasing the local train steps on the \method~ performance. As shown in \citep{FedProx}, a reduction in the communication costs can be achieved by increasing $E$ at the expense of local models convergence, which can substantially affect the aggregation process. Thus, with this analysis, we aim to \textit{understand whether \method~ models can retain their convergence rate when multiple local train steps are performed across clients' data to reduce the communication costs.} To this end, we perform experiments with various labeled percentages for 50 federated rounds ($R$=50) and $N$=15, while varying $E$ from 1 to 4. From the results shown in Figure~\ref{fig:robustness_E}, we note that \method~ can effectively utilize the unlabeled instances, when the available labeled subset exceeds 3\%, to avoid possible local models' divergence, resulting in a highly accurate aggregated (or global) model. However, for $L$=3\%, we notice a declining trend as $E$ increases, which could be originated from two reasons. Since the labeled data are scant for $L$=3\%, the downwards trend on \method~ performance could be caused due to over-fitting, as the local models are extensively trained on a tiny labeled subset, when $E$ increases. In addition, the absence of such a trend in higher label availability rates suggests that a sufficient amount of labeled data might be required for \method~ local models to converge. As \method~ uses unlabeled instances predictions to retrain the local models further, any faulty pseudo-labeled samples participating in the retraining step will increasingly intensify local models' divergence as $E$ rises. Besides providing additional labels to avoid local models divergence, this behavior could be regulated by adjusting the confidence threshold of the predictions, $\tau$, to a higher value so that any initial faulty pseudo-labels would participate in the local SGD steps are discarded. \par

\begin{figure}[h]
    \centering \small
    \includegraphics[width=0.75\textwidth]{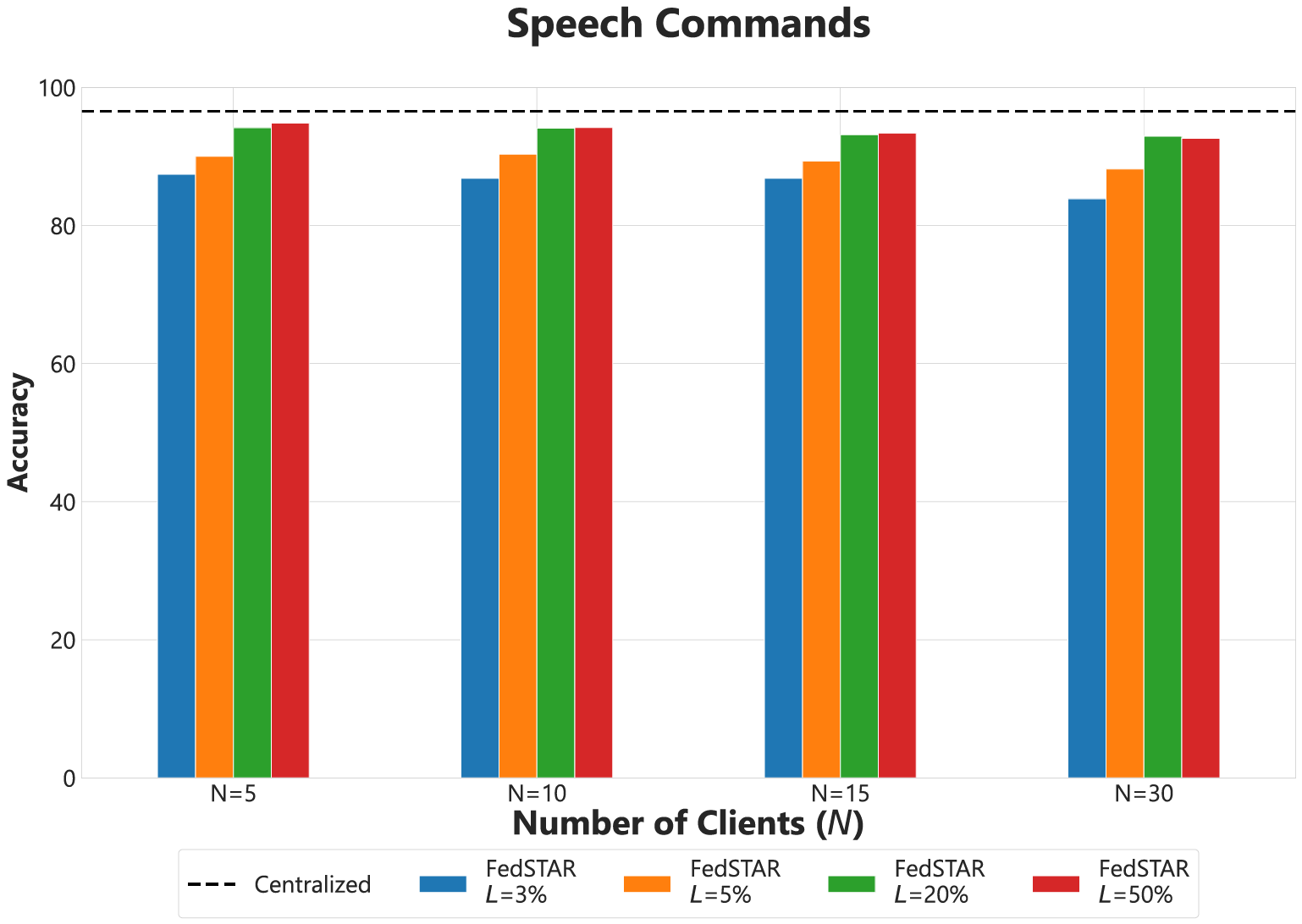}
    \caption{Evaluation of \method~ performance under varying number of clients. Federated parameters are set to $\sigma$=25\%, $\beta=0.5$, $R$=100, $q$=80\% and $E$=1. Average accuracy over 3 distinct trials is reported.} \label{fig:robustness_N}
\end{figure}

\textbf{Varying number of clients:} The number of clients is an important factor in the FL procedure, as it can have a significant impact on the data distribution, which has shown to affect the global model's generalization \citep{NonIID, OpenChallenges}. In particular, introducing additional clients to FL, the class distributions across clients can become highly skewed, as the data partitioning process is random. With this ablation study, we aim to answer \textit{whether \method~ can retain the same level of effectiveness when the number of clients (and thus the non-i.i.d-ness of the class distribution) grows.} To this end, we present the performance of \method~ on Speech Commands dataset, when we vary the number of clients from 5 up to 30, while setting $q$=80\%, $\sigma$=25\%, $R$=100\% and $E$=1. The findings are shown in Figure~\ref{fig:robustness_N}, where we note that the number of clients has a relatively low impact on \method~ ability to utilize the available unlabeled audio data, as \method~ models' performance follows a constant upward trend while we provide additional labels for any given $N$. In particular, comparing the results for $N$=5 and $N$=30, we observe that the \method~ models' accuracy is notably close, especially for $L$>3\%. This is in constant with fully supervised FL performance, as presented in Table~\ref{tab:baseline}, where models' performance can drop more than 2\% when varying $N$ for the same dataset. Finally, it is important to note that \method~ model's performance is close to the centralized baseline for both $L$>20 and $L$=50; thus, no noticeable improvement appears with the introduction of additional labels samples.

\textbf{Varying class distribution across clients:} Apart from the number of clients, the preferences of each client can substantially affect the nature of clients' data distribution. For example, in a music tagging scenario, the type and quantity of data residing on a device are directly correlated to both user's preference of a specific genre of music and the time user's dedicated to the application. Such challenges introduce a highly non-i.i.d. data distribution, both in terms of labels distribution and quantity of data per client. Therefore, in this analysis, we aim to \textit{understand the effect of highly non-i.i.d. distributions, both in terms of labels and data quantity distributions, on the effectiveness of \method~ to utilize on-device unlabeled data.} To this end, we execute experiments on the Speech Commands dataset with $N$=15, $q$=80\% and $E$=1 for $R$=100, in which the partitioning of labeled data on clients followed a defined class availability distribution. We utilize a uniform distribution with a mean value of $\mu$=3 and fluctuating variance $\sigma_{c}$ from 0\% to 50\% as our class availability distribution across clients. Since the total number of classes in the Speech Commands dataset is 12, we choose $\mu$=3 for clients to access only a few labeled samples per class (on average, three classes). Thus, the on-device labeled data distribution resembles a realistic non-i.i.d. distribution. The client's preferences can affect both the type and the number of labeled samples described earlier for a music tagging application. It is important to note that the splitting of the unlabeled subset on clients followed a random distribution, with no assumption being made to distribute the label. Consequently, clients might have labeled samples from a specific subset of classes, yet unlabeled instances from all classes could be available. Such data distributions are frequent in pragmatic applications, where the domain knowledge is missing to perform the annotation process appropriately for all classes. For a rigorous evaluation, we perform identical experiments in terms of on-device labeled samples availability under fully supervised federated settings, where the unlabeled dataset remained unexploited. \par

From the results introduced in Table~\ref{tab:c_dist}, we note that \method~ can effectively exploit the available on-device unlabeled instances to learn an accurate audio model under highly non-i.i.d. distributions. Comparing the \method~ performance with that of a fully supervised FL counterpart, we notice a substantial improvement in accuracy in most cases. In particular, for the case of $L\le$3\%, \method~ utilized on-device unlabeled examples to effectively train an audio model, whereas FL was unable to learn under such highly non-i.i.d. settings adequately. Additionally, we observe that the obtained accuracy gap across three distinct \method~ models (with $\sigma_{c}$ of 0, 25, and 50 percent) for a given $L$ is no larger than 4.8\%. This behavior suggests that \method~ can maintain nearly the same level of effectiveness in exploiting on-device unlabeled data, irrespective of the skewness of data distribution on clients' end. Consequently, \method~ could be an effective solution to train an audio model under different federated settings, where the labeled data across clients experience a class distribution skewness and large-scale unlabeled audio samples from all classes are readily available on clients' devices. \par

\begin{table*}[t]
\centering \small
\caption{Performance evaluation of method~ against variation of class availability across clients. Class distribution has mean $\mu$=3 and variance $\sigma_{c}$. Average accuracy over 3 distinct runs is reported on Speech Commands. Detailed results are given in Table~\ref{tab:c_dist_std} of the Appendix. Federated parameters are set to $\beta=0.5$, $R$=100, $N$=15, $q$=80\% and $E$=1.} \label{tab:c_dist}
\resizebox{\textwidth}{!}{%
    \begin{tabular}{lc cccccccc}
        \toprule
        \multicolumn{2}{c}{\multirow{2}{*}{\begin{tabular}[c]{@{}c@{}}\textbf{Class Distribution}\\ \textbf{Characteristics}\end{tabular}}}
        & \multicolumn{4}{c}{\textbf{Supervised (Federated)}} 
        & \multicolumn{4}{c}{\textbf{FedSTAR}} \\ 
        \cmidrule[0.5pt](rl){3-6} \cmidrule[0.5pt](rl){7-10} 
        &   & $L=3\%$ & $L=5\%$ & $L=20\%$ & $L=50\%$ 
            & $L=3\%$ & $L=5\%$ & $L=20\%$ & $L=50\%$ \\ 
    
        \midrule[0.5pt]
        \multirow{3}{*}{$\mu$=3} & $\sigma_{c}$=0 \%
                    & 9.83              & 32.63             & 80.22             & 82.40 
                    & \textbf{79.08}    & \textbf{79.62}    & \textbf{87.01}    & \textbf{83.14} \\

        & $\sigma_{c}$=25\%
                    & 10.54             & 23.97             & 75.41             & 83.61 
                    & \textbf{79.05}    & \textbf{84.15}    & \textbf{86.52}    & \textbf{85.05} \\

        & $\sigma_{c}$=50\%     
                    & 8.44              & 24.25             & 73.93             & 84.41
                    & \textbf{78.14}    & \textbf{81.88}    & \textbf{84.56}    & \textbf{84.55} \\
        \bottomrule
    \end{tabular}%
}
\end{table*}

\subsubsection{\textbf{Assessment of utilizing self-supervised learning for model pre-training to improve training convergence of~\method.}} \label{sec:ssl_train} 
\begin{figure}[t]
    \centering \small
    \subfigure[Accuracy obtained after 10 rounds for $N$=15 on Speech Commands.]{ \centering \label{fig:spcm_bar} \includegraphics[width=0.48\textwidth]{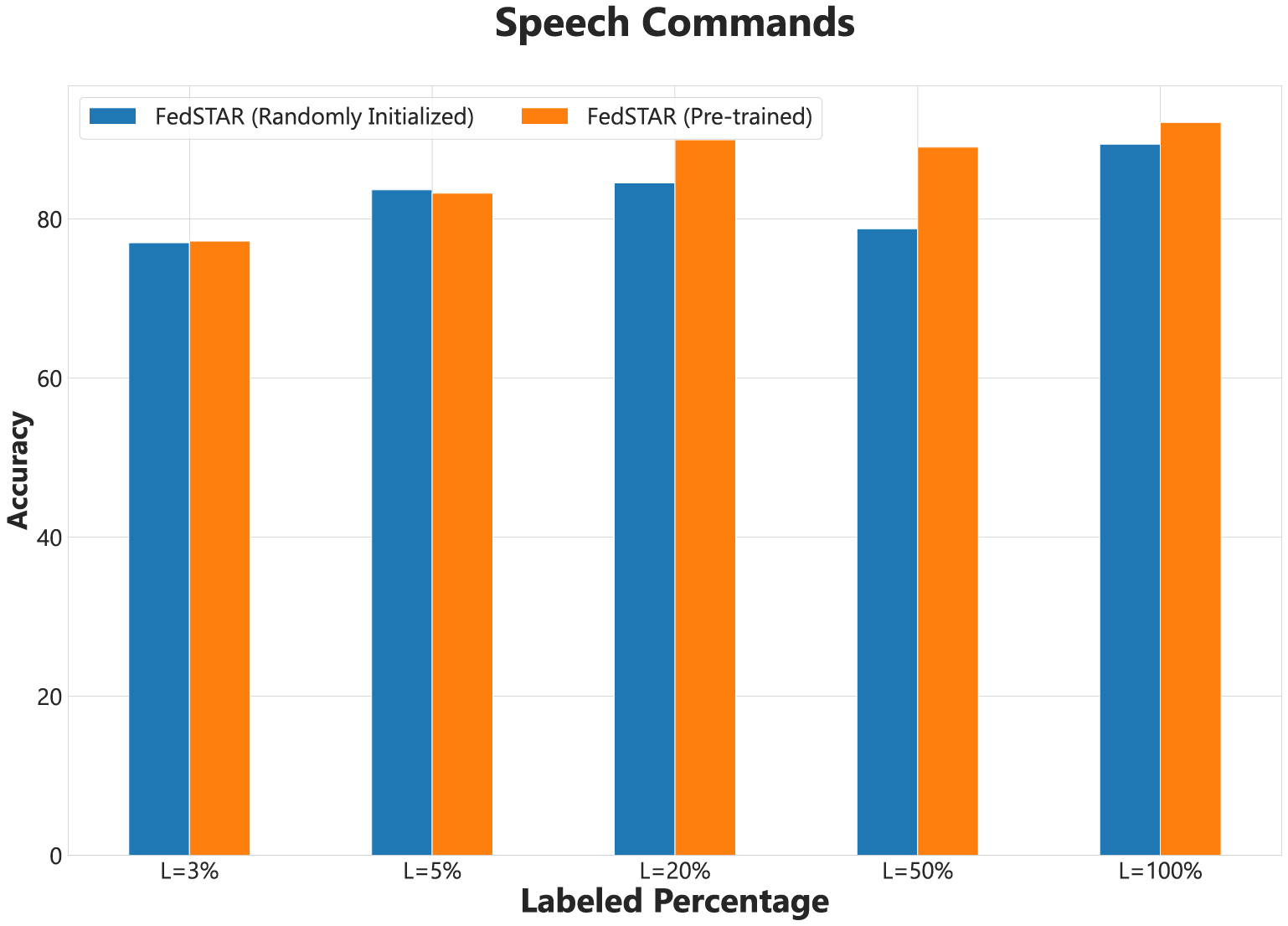}} \hfill
    \subfigure[Train loss comparison for $N$=15 on Speech Commands.]{ \centering \label{fig:spcm_line} \includegraphics[width=0.48\textwidth]{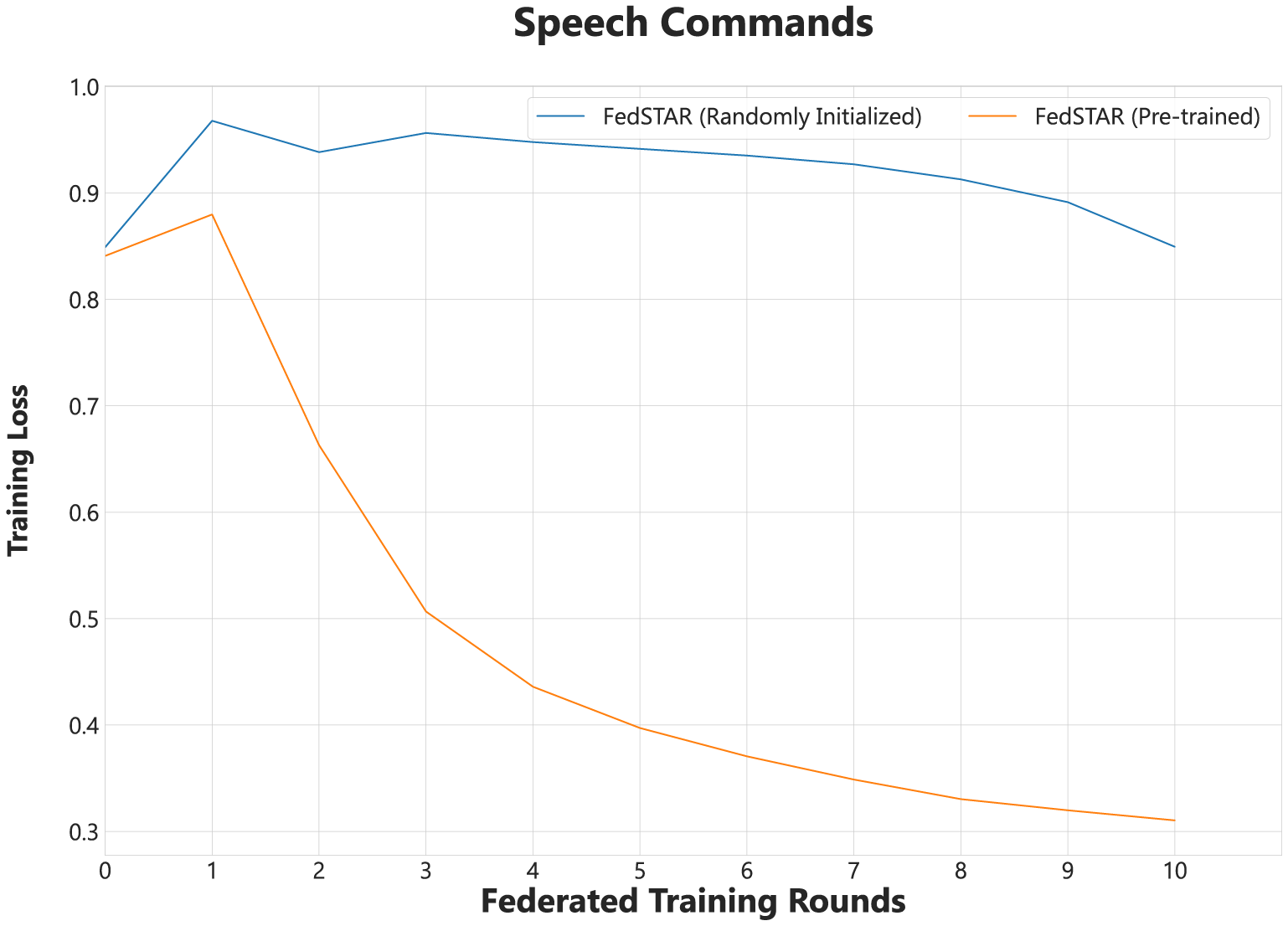}}
    
    \subfigure[Accuracy obtained after 10 rounds for $N$=15 on Ambient Context.]{\centering \label{fig:abcx_bar} \includegraphics[width=0.48\textwidth]{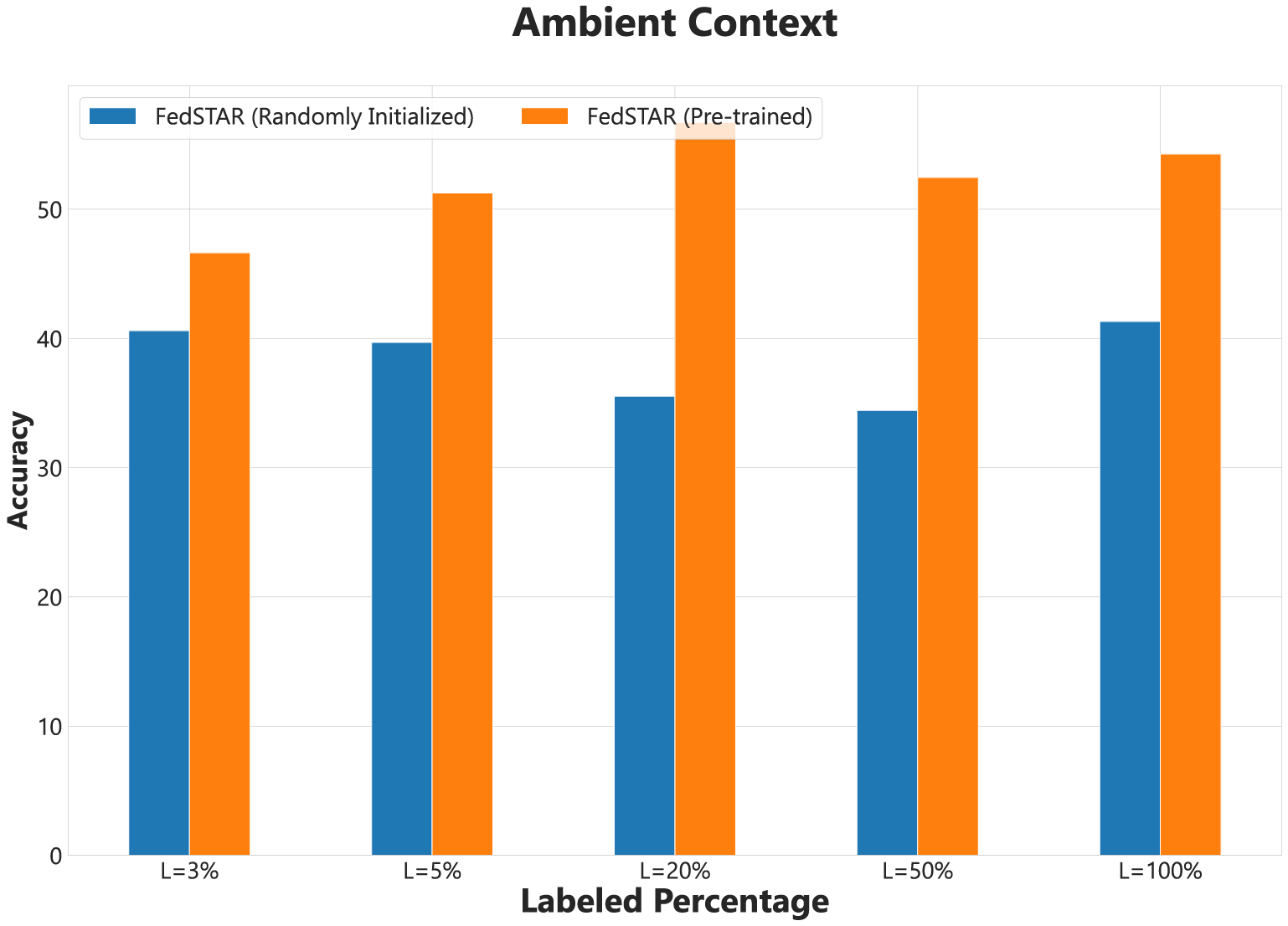}} \hfill
    \subfigure[Train loss comparison for $N$=15 on Ambient Context.]{\centering \label{fig:abcx_line} \includegraphics[width=0.48\textwidth]{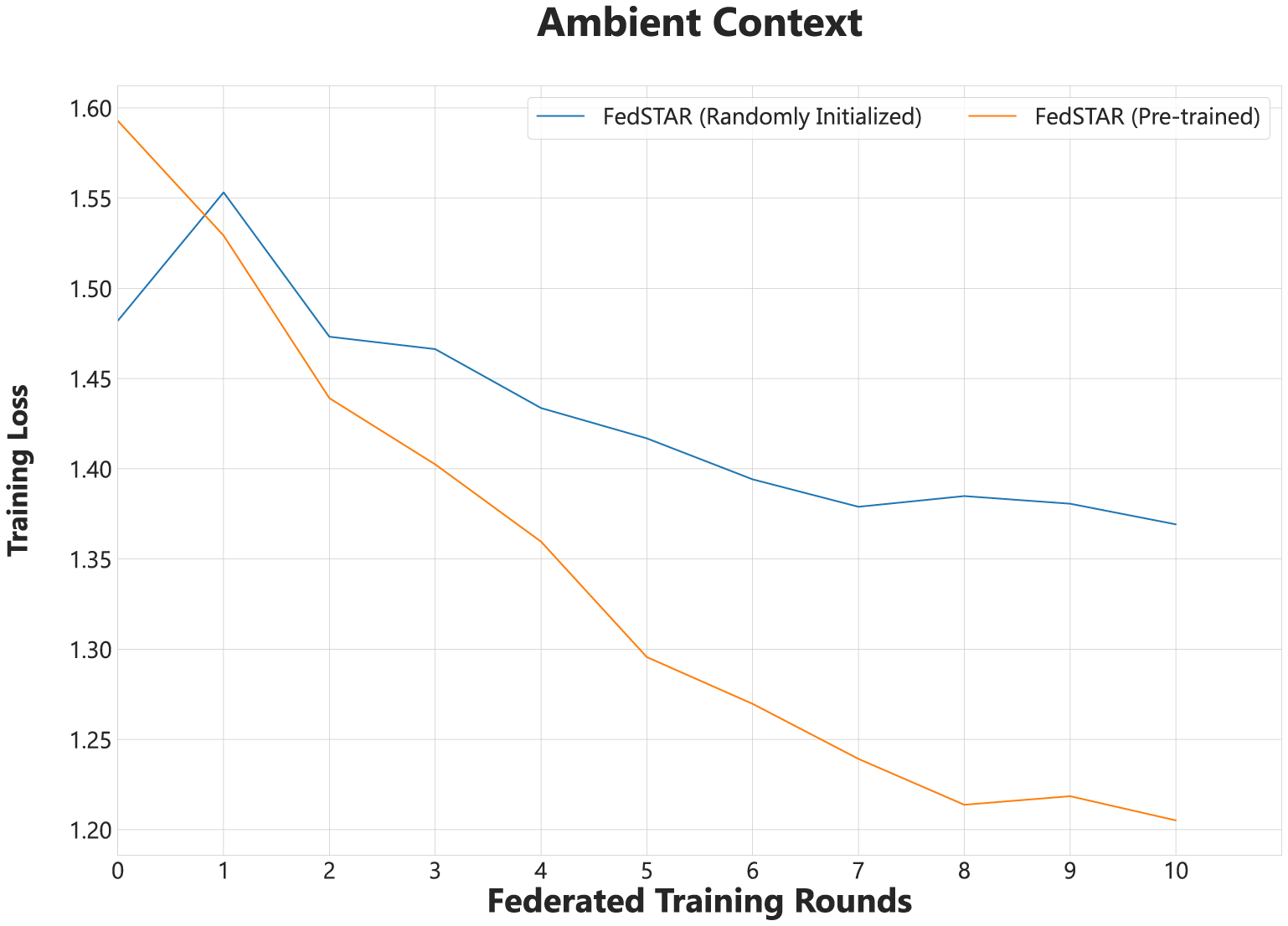}}
    
    \subfigure[Accuracy obtained after 10 rounds for $N$=15 on VoxForge.]{\centering \label{fig:vf_bar} \includegraphics[width=0.48\textwidth]{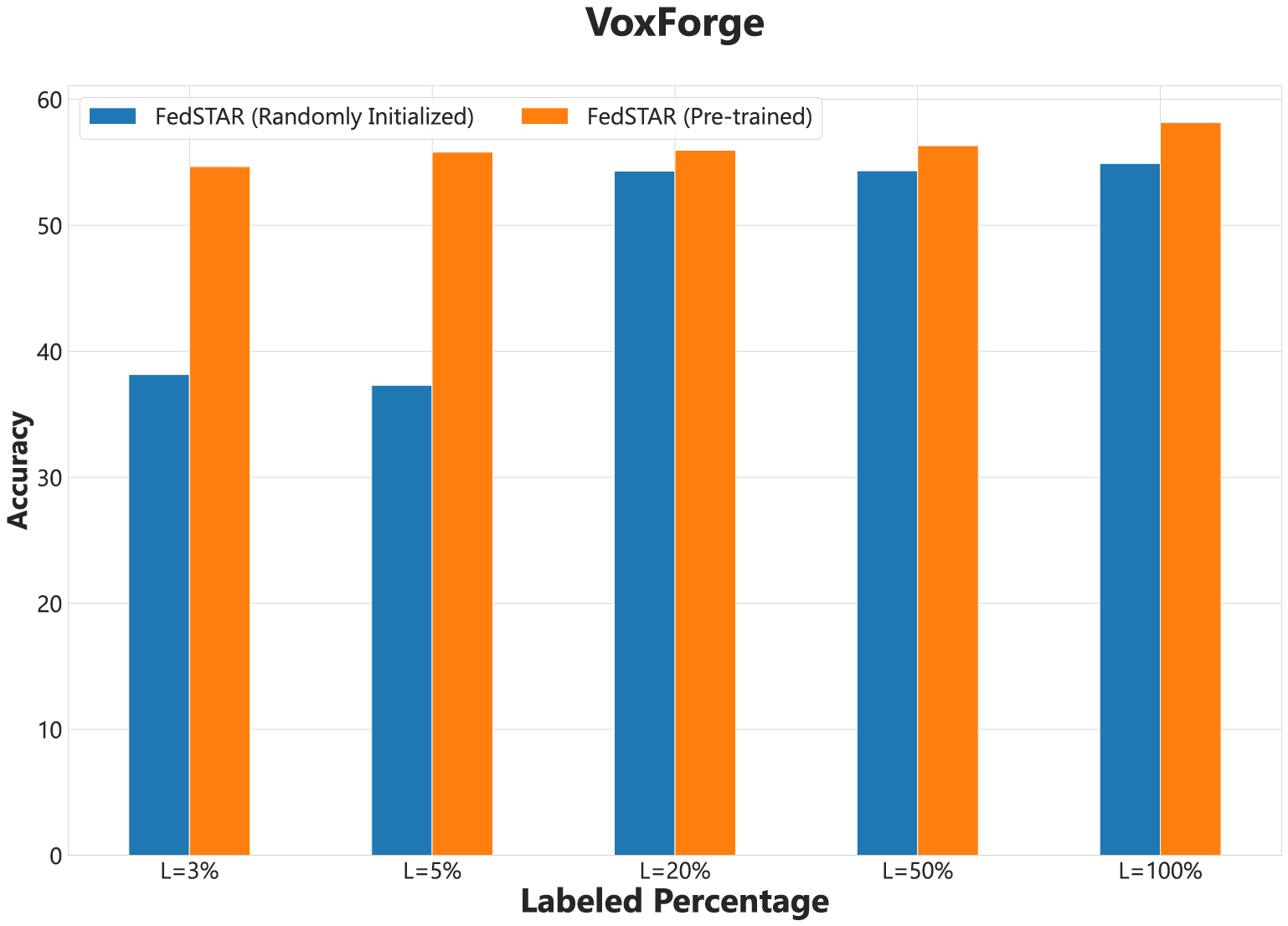}} \hfill
    \subfigure[Train loss comparison for $N$=15 on VoxForge.]{\centering \label{fig:vf_line} \includegraphics[width=0.48\textwidth]{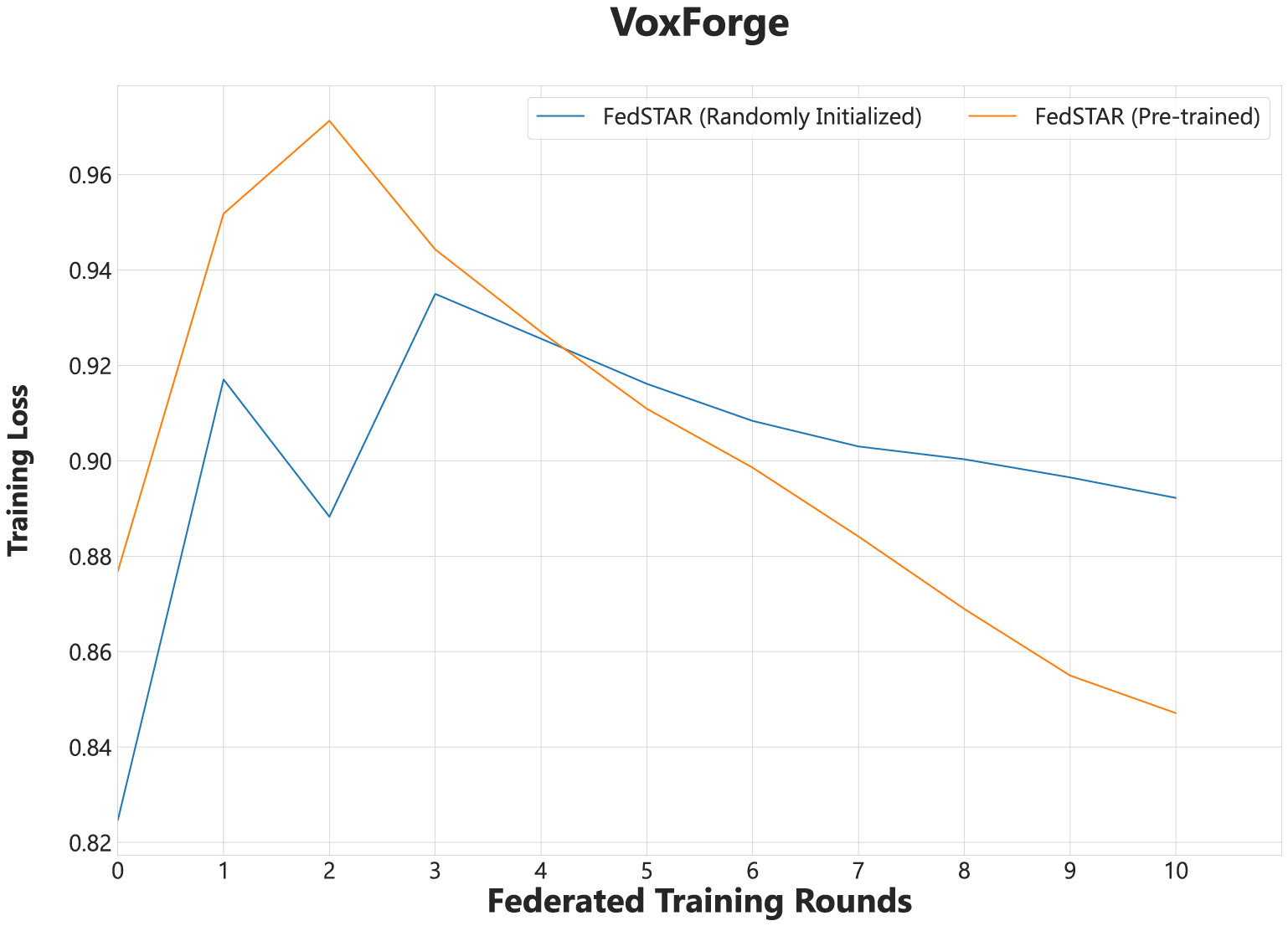}}

    \caption{Self-supervised learning improves training convergence in federated setting. Federated parameters are set to $q$=80\%, $\sigma$=25\%, $\beta$=0.5, $E$=1. Average accuracy on testset over three distinct trials is reported.} \label{fig:PSL_SSL}
\end{figure}

Our proposed self-training federated learning approach attains high performance on different audio recognition tasks by utilizing unlabeled data available on clients' end. However, in reality, a large volume of unlabeled instances from a different task or distribution might also be available on the centralized server. As servers often possess the computational power to effectively pre-train a model on a massive unlabeled dataset, a natural question arises, \textit{whether leveraging self-supervised learning to pre-train a model as initialization for \method~ could improve the training convergence in federated settings with fewer rounds}. To this end, we perform experiments on all three datasets with $N$=15 while using a model trained with a self-supervised pre-training strategy introduced in Section~\ref{sec:pre_train}. We compare the obtained accuracy after ten rounds of training ($R$=10) when utilizing a self-supervised pre-trained model as an initial starting point for \method~ in contrast with a randomly initialized \method~ model trained for the same number of federated rounds. For a more rigorous evaluation, we vary labels availability from L=3\% up to L=100\% across all our datasets. The findings are presented in~\Cref{fig:spcm_bar,fig:abcx_bar,fig:vf_bar}, where the average accuracy over three distinct trials is reported. Furthermore, the average train loss for the case of $L$=50\% in the first 10 federated rounds ($R$=10) is also reported in~\Cref{fig:spcm_line,fig:abcx_line,fig:vf_line}. We choose to report the average train loss for the case of $L$=50\% since we previously observed from Table~\ref{tab:results} that \method~ models might require additional rounds to utilize unlabeled data effectively. Thus, we can demonstrate that utilizing a pre-trained model as initialization for \method~ can significantly boost training convergence. \par

From~\Cref{fig:spcm_bar,fig:abcx_bar,fig:vf_bar}, we note that the utilization of a pre-trained model leads to higher accuracy within $10$ rounds in almost all cases, suggesting that it was able to perform finer pseudo-labels predictions and accelerate the model's convergence. In particular, for the Ambient Context dataset, where the amount of available labeled instances per client is tiny (approximately 13 labeled samples per client for $L$=5\% and $N$=15), we observe a substantial difference between the pre-trained and randomly initialized \method~ approaches. This behavior suggests that in cases where the amount of labels is exceedingly sparse, utilizing self-supervised learning via model pre-training can significantly shorten the federated rounds needed for convergence. The beneficial role of the model pre-training on \method~ can be observed in the train loss gap between the pre-trained and the randomly initialized versions of \method~ after ten rounds, presented in ~\Cref{fig:spcm_line,fig:abcx_line,fig:vf_line}. \par

\subsubsection{\textbf{Effectiveness of \method~ under varying amount of unlabeled data.}} \label{sec:ssl_u} As of now, we have assumed that unlabeled data is largely available across clients. However, it is intriguing to investigate the scenario where both the amount of labeled and unlabeled data varies. In this way, we could simulate two pragmatic scenarios: First, an abundant volume of unlabeled instances generated by clients devices (e.g., numerous IoT devices constantly monitoring the surrounding environment); and second, relatively small amount of unlabeled audio samples available (e.g., medical audio examples, where, both obtaining and labeling data is expensive). In addition, the restriction of available unlabeled on-device data could be originated from the often-limited storage capabilities of devices participating in the distributed machine learning paradigms. Thus, we aim to understand \textit{the effect of unlabeled data availability on the \method~ efficiency to improve FL models' performance as well as the impact of utilizing pre-trained model with self-supervised}. Consequently, we perform experiments on the Speech Commands dataset with $N$=15, while varying the labeled subset from 3\% up to 50\% and the unlabeled dataset from 20\% up to 100\%. The obtained accuracy scores for both pre-trained and randomly initialized \method~ models are presented in Table~\ref{tab:vary_u}. \par

\begin{table*}[t]
\centering \small
\caption{Performance evaluation of \method~ when varying both labeled and unlabeled datasets. 
Average accuracy over 3 distinct runs is reported on Speech Commands. 
Detailed results are given in Table~\ref{tab:vary_u_std} of the Appendix. 
Federated parameters are set to $q$=80\%, $\sigma$=25\%, $\beta$=0.5, $R$=100, $E$=1, $N$=15.} \label{tab:vary_u}
\begin{tabular}{l cccc cccc}
    \toprule
    \multicolumn{1}{c}{\multirow{2}{*}{\begin{tabular}[c]{@{}l@{}}\textbf{Labeled}\\\textbf{Percentage}\end{tabular}}} & \multicolumn{4}{c}{\textbf{FedSTAR (Randomly Initialized)}} & \multicolumn{4}{c}{\textbf{FedSTAR (SSL Pretrained)}} \\
    \cmidrule[0.5pt](rl){2-5} \cmidrule[0.5pt](rl){6-9}

    & $U$=20\% & $U$=50\% & $U$=80\% & $U$=100\% & $U$=20\% & $U$=50\% & $U$=80\% & $U$=100\% \\

    \midrule[0.5pt]
    $L$=3\%   & 84.13 & \textbf{85.40} & \textbf{86.63} & \textbf{86.82}    & \textbf{84.52} & 85.17	        & 85.43	& 86.46 \\
    $L$=5\%   & 87.47 & \textbf{88.52} & \textbf{88.90} & \textbf{89.33}    & \textbf{88.07} & 88.28	        & 87.73	& 88.98 \\
    $L$=20\%  & 90.06 & 92.24 & 93.07 & 93.15                               & \textbf{92.44} & \textbf{93.67}	& \textbf{93.98}	& \textbf{94.13} \\
    $L$=50\%  & 87.76 & 92.26 & 94.18 & 93.38                               & \textbf{90.70} & \textbf{93.83}	& \textbf{94.76}	& \textbf{95.54} \\
    \bottomrule
\end{tabular}%
\end{table*}

As we see in Table~\ref{tab:vary_u}, the availability of unlabeled data can affect the \method~ models performance. In particular, when the amount of both labeled and unlabeled instances is limited ($L\le$5 and $U\le$50), the obtained accuracy for both the randomly initialized and the pre-trained \method~ models is similar. However, as the amount of labeled data increases ($L$>5), we notice a performance improvement of the pre-trained over the standard \method~ approach, in case of $L$=50 and $U$=20 results an accuracy difference of 3\%. This behavior suggests that the utilization of on-device unlabeled data for the pre-trained \method~ is superior to that of the randomly initialized \method~ when sufficient labeled samples are provided. In addition, comparing results of both \method~ approaches in Table~\ref{tab:vary_u} with those for supervised federated with $N$=15 in Table~\ref{tab:results}, we observe an accuracy improvement for both pre-trained and randomly initialized \method~ models over their supervised counterparts for $L$<20. For higher labels availability ($L$>20), the performance gap between the pre-trained \method~ models and the fully-supervised federated alternatives is still prevalent, even for small volumes of unlabeled data ($U$=20). From this, we can deduce that \method~ can effectively utilize unlabeled instances to improve the performance of audio models, even when the availability of on-device unlabeled samples is insufficient.

\section{Conclusions and Future Work}
We study the pragmatic problem of semi-supervised federated learning for audio recognition tasks. In the distributed scenario, clients' well-annotated audio examples are deficient due to the prohibitive cost of annotation. Users with little to no incentives to label their data, and notably for various important tasks, the domain knowledge is missing to perform the annotation process appropriately. Conversely, large-scale unlabeled audio data are readily available on clients' devices. To address the lack of labeled data for learning on-device models, we present a novel self-training strategy based on pseudo-labeling to exploit on-device unlabeled audio data and boost the generalization of models trained in federated settings. Despite its simplicity, we demonstrate that our approach, \method, is highly feasible for semi-supervised learning on various audio recognition tasks within different federated settings and labels availability. We exhaustively evaluate \method~ on several publicly available datasets while comparing its performance with fully-supervised federated and traditional centralized counterparts. The models' accuracy we achieve is consistently superior to fully supervised federated settings under the same labels availability. In many cases, \method~ results are comparable to fully-supervised federated settings, where the complete dataset with labels was utilized. Furthermore, \method~ can retain the same level of effectiveness on utilizing unlabeled instances, irrespective of the amount of labels available on clients. In addition, \method~ can significantly improve the model's performance in settings where on-device labeled samples from only a subset of classes are present, while the unlabeled instances contain examples from all classes. By utilizing on-device unlabeled samples from all classes, the data distribution across devices becomes more uniform; thus, the local models' learning objectives converge. This non-i.i.d data distribution setting is frequent in pragmatic scenarios, where the expertise is missing to annotate samples from all available classes, e.g., physiological signals in the medical domain.  Finally, we demonstrate that self-supervised pre-trained models can significantly improve training convergence in federated settings with fewer rounds when used as model initialization for federated training instead of randomly initialized weights. 

Despite the wide applicability, as \method~is based on self-training, it is still relying on a few well-annotated samples across all devices to properly exploit any additional unlabeled data. Without such labeled samples, the utilization of unlabeled samples though \method~might bring undesirable results. In reality, however, such a limitation can be lifted by requesting from users to annotate $2-3$ samples, which are inexpensive to acquire. With the number of devices in a FL network usually ranging from hundreds to even thousands of devices, this process will provide a sufficient labeled subset. This can be used to train model in conjunction with the massively available unlabeled data using~\method~to acquire a highly accurate model. Furthermore, inherent noise, which originate from the audio signal is an additional challenge that can limit the applicability of~\method in real-life applications. Depending on the type and the amount of noise, this could affect the performance of~\method~ making the exploitation of unlabeled samples counter-productive. In such cases, there are a range of methods available that can be introduced as a prepossessing step in the learning procedure to mitigate or denoise the signal with minimal effort.

In this work, we provided a federated self-training scheme to learn audio recognition models through a few on-device labeled audio data. In the Internet of Things era, this approach could be employed in a variety of applications, such as home automation, autonomous driving, the healthcare domain, and smart wearable technologies. In particular, we believe that federated self-training is of immense value for learning generalizable audio models in settings, where, labeled data are challenging to acquire. However, unlabeled data are available in vast quantities. We hope that the presented perspective of federated self-training inspires the development of additional approaches, specifically those combining semi-supervised learning and federated learning in an asynchronous fashion. Likewise, combining federated self-training with appropriate client selection techniques is another crucial area of improvement that will further improve the performance of deep models in federated learning scenarios. Finally, evaluation in a real-world setting (i.e., federate learning involving real devices) is of major importance to further understand the aspects that require improvements concerning statistical and system heterogeneities, energy and, labeled data requirements in the federated learning setting. \par 

\begin{acks}
Various icons used in the figures are created by Teewara Soontorn, Becris, Atif Arshad, Graphic Tigers, Stefan Traistaru, and Andrejs Kirma from the Noun Project.
\end{acks}

\bibliographystyle{ACM-Reference-Format}
\bibliography{main}


\begin{thebibliography}{46}


\ifx \showCODEN    \undefined \def \showCODEN     #1{\unskip}     \fi
\ifx \showDOI      \undefined \def \showDOI       #1{#1}\fi
\ifx \showISBNx    \undefined \def \showISBNx     #1{\unskip}     \fi
\ifx \showISBNxiii \undefined \def \showISBNxiii  #1{\unskip}     \fi
\ifx \showISSN     \undefined \def \showISSN      #1{\unskip}     \fi
\ifx \showLCCN     \undefined \def \showLCCN      #1{\unskip}     \fi
\ifx \shownote     \undefined \def \shownote      #1{#1}          \fi
\ifx \showarticletitle \undefined \def \showarticletitle #1{#1}   \fi
\ifx \showURL      \undefined \def \showURL       {\relax}        \fi
\providecommand\bibfield[2]{#2}
\providecommand\bibinfo[2]{#2}
\providecommand\natexlab[1]{#1}
\providecommand\showeprint[2][]{arXiv:#2}

\bibitem[\protect\citeauthoryear{Arazo, Ortego, Albert, O’Connor, and
  McGuinness}{Arazo et~al\mbox{.}}{2020}]%
        {DeepPSL}
\bibfield{author}{\bibinfo{person}{Eric Arazo}, \bibinfo{person}{Diego Ortego},
  \bibinfo{person}{Paul Albert}, \bibinfo{person}{Noel~E. O’Connor}, {and}
  \bibinfo{person}{Kevin McGuinness}.} \bibinfo{year}{2020}\natexlab{}.
\newblock \showarticletitle{Pseudo-Labeling and Confirmation Bias in Deep
  Semi-Supervised Learning}. In \bibinfo{booktitle}{\emph{2020 International
  Joint Conference on Neural Networks (IJCNN)}}. \bibinfo{pages}{1--8}.
\newblock
\urldef\tempurl%
\url{https://doi.org/10.1109/IJCNN48605.2020.9207304}
\showDOI{\tempurl}


\bibitem[\protect\citeauthoryear{Berthelot, Carlini, Goodfellow, Papernot,
  Oliver, and Raffel}{Berthelot et~al\mbox{.}}{2019}]%
        {MixMatch}
\bibfield{author}{\bibinfo{person}{David Berthelot}, \bibinfo{person}{Nicholas
  Carlini}, \bibinfo{person}{Ian Goodfellow}, \bibinfo{person}{Nicolas
  Papernot}, \bibinfo{person}{Avital Oliver}, {and} \bibinfo{person}{Colin
  Raffel}.} \bibinfo{year}{2019}\natexlab{}.
\newblock \bibinfo{title}{MixMatch: A Holistic Approach to Semi-Supervised
  Learning}.
\newblock
\newblock
\showeprint[arxiv]{1905.02249}~[cs.LG]


\bibitem[\protect\citeauthoryear{Beutel, Topal, Mathur, Qiu, Parcollet, and
  Lane}{Beutel et~al\mbox{.}}{2020}]%
        {Flower}
\bibfield{author}{\bibinfo{person}{Daniel~J Beutel}, \bibinfo{person}{Taner
  Topal}, \bibinfo{person}{Akhil Mathur}, \bibinfo{person}{Xinchi Qiu},
  \bibinfo{person}{Titouan Parcollet}, {and} \bibinfo{person}{Nicholas~D
  Lane}.} \bibinfo{year}{2020}\natexlab{}.
\newblock \showarticletitle{Flower: A Friendly Federated Learning Research
  Framework}.
\newblock \bibinfo{journal}{\emph{arXiv preprint arXiv:2007.14390}}
  (\bibinfo{year}{2020}).
\newblock


\bibitem[\protect\citeauthoryear{Chan, Rea, Gollakota, and Sunshine}{Chan
  et~al\mbox{.}}{2019}]%
        {chan2019contactless}
\bibfield{author}{\bibinfo{person}{Justin Chan}, \bibinfo{person}{Thomas Rea},
  \bibinfo{person}{Shyamnath Gollakota}, {and} \bibinfo{person}{Jacob~E
  Sunshine}.} \bibinfo{year}{2019}\natexlab{}.
\newblock \showarticletitle{Contactless cardiac arrest detection using smart
  devices}.
\newblock \bibinfo{journal}{\emph{NPJ digital medicine}} \bibinfo{volume}{2},
  \bibinfo{number}{1} (\bibinfo{year}{2019}), \bibinfo{pages}{1--8}.
\newblock


\bibitem[\protect\citeauthoryear{Foggia, Petkov, Saggese, Strisciuglio, and
  Vento}{Foggia et~al\mbox{.}}{2016}]%
        {RoadSurveillance}
\bibfield{author}{\bibinfo{person}{Pasquale Foggia}, \bibinfo{person}{Nicolai
  Petkov}, \bibinfo{person}{Alessia Saggese}, \bibinfo{person}{Nicola
  Strisciuglio}, {and} \bibinfo{person}{Mario Vento}.}
  \bibinfo{year}{2016}\natexlab{}.
\newblock \showarticletitle{Audio Surveillance of Roads: A System for Detecting
  Anomalous Sounds}.
\newblock \bibinfo{journal}{\emph{IEEE Transactions on Intelligent
  Transportation Systems}} \bibinfo{volume}{17}, \bibinfo{number}{1}
  (\bibinfo{year}{2016}), \bibinfo{pages}{279--288}.
\newblock
\urldef\tempurl%
\url{https://doi.org/10.1109/TITS.2015.2470216}
\showDOI{\tempurl}


\bibitem[\protect\citeauthoryear{Fonseca, Favory, Pons, Font, and
  Serra}{Fonseca et~al\mbox{.}}{2020}]%
        {fonseca2020fsd50k}
\bibfield{author}{\bibinfo{person}{Eduardo Fonseca}, \bibinfo{person}{Xavier
  Favory}, \bibinfo{person}{Jordi Pons}, \bibinfo{person}{Frederic Font}, {and}
  \bibinfo{person}{Xavier Serra}.} \bibinfo{year}{2020}\natexlab{}.
\newblock \showarticletitle{FSD50k: an open dataset of human-labeled sound
  events}.
\newblock \bibinfo{journal}{\emph{arXiv preprint arXiv:2010.00475}}
  (\bibinfo{year}{2020}).
\newblock


\bibitem[\protect\citeauthoryear{Gao, Parcollet, Fernandez-Marques, de~Gusmao,
  Beutel, and Lane}{Gao et~al\mbox{.}}{2021}]%
        {TrainEESR}
\bibfield{author}{\bibinfo{person}{Yan Gao}, \bibinfo{person}{Titouan
  Parcollet}, \bibinfo{person}{Javier Fernandez-Marques},
  \bibinfo{person}{Pedro P.~B. de Gusmao}, \bibinfo{person}{Daniel~J. Beutel},
  {and} \bibinfo{person}{Nicholas~D. Lane}.} \bibinfo{year}{2021}\natexlab{}.
\newblock \bibinfo{title}{End-to-End Speech Recognition from Federated Acoustic
  Models}.
\newblock
\newblock
\showeprint[arxiv]{2104.14297}~[cs.SD]


\bibitem[\protect\citeauthoryear{Grandvalet and Bengio}{Grandvalet and
  Bengio}{2004}]%
        {Entropy}
\bibfield{author}{\bibinfo{person}{Yves Grandvalet} {and}
  \bibinfo{person}{Yoshua Bengio}.} \bibinfo{year}{2004}\natexlab{}.
\newblock \showarticletitle{Semi-Supervised Learning by Entropy Minimization}.
  In \bibinfo{booktitle}{\emph{Proceedings of the 17th International Conference
  on Neural Information Processing Systems}} (Vancouver, British Columbia,
  Canada) \emph{(\bibinfo{series}{NIPS'04})}. \bibinfo{publisher}{MIT Press},
  \bibinfo{address}{Cambridge, MA, USA}, \bibinfo{pages}{529–536}.
\newblock


\bibitem[\protect\citeauthoryear{Hard, Partridge, Nguyen, Subrahmanya, Shah,
  Zhu, Moreno, and Mathews}{Hard et~al\mbox{.}}{2020}]%
        {KeywordSpottingNonIID}
\bibfield{author}{\bibinfo{person}{Andrew Hard}, \bibinfo{person}{Kurt
  Partridge}, \bibinfo{person}{Cameron Nguyen}, \bibinfo{person}{Niranjan
  Subrahmanya}, \bibinfo{person}{Aishanee Shah}, \bibinfo{person}{Pai Zhu},
  \bibinfo{person}{Ignacio~Lopez Moreno}, {and} \bibinfo{person}{Rajiv
  Mathews}.} \bibinfo{year}{2020}\natexlab{}.
\newblock \bibinfo{title}{Training Keyword Spotting Models on Non-IID Data with
  Federated Learning}.
\newblock
\newblock
\showeprint[arxiv]{2005.10406}~[eess.AS]


\bibitem[\protect\citeauthoryear{Hard, Rao, Mathews, Beaufays, Augenstein,
  Eichner, Kiddon, and Ramage}{Hard et~al\mbox{.}}{2018}]%
        {KeyboardPrediction}
\bibfield{author}{\bibinfo{person}{Andrew Hard}, \bibinfo{person}{Kanishka
  Rao}, \bibinfo{person}{Rajiv Mathews}, \bibinfo{person}{Fran{\c{c}}oise
  Beaufays}, \bibinfo{person}{Sean Augenstein}, \bibinfo{person}{Hubert
  Eichner}, \bibinfo{person}{Chlo{\'{e}} Kiddon}, {and} \bibinfo{person}{Daniel
  Ramage}.} \bibinfo{year}{2018}\natexlab{}.
\newblock \showarticletitle{Federated Learning for Mobile Keyboard Prediction}.
\newblock \bibinfo{journal}{\emph{CoRR}}  \bibinfo{volume}{abs/1811.03604}
  (\bibinfo{year}{2018}).
\newblock
\showeprint[arxiv]{1811.03604}
\urldef\tempurl%
\url{http://arxiv.org/abs/1811.03604}
\showURL{%
\tempurl}


\bibitem[\protect\citeauthoryear{Hinton, Vinyals, and Dean}{Hinton
  et~al\mbox{.}}{2015}]%
        {DistillingPSL}
\bibfield{author}{\bibinfo{person}{Geoffrey Hinton}, \bibinfo{person}{Oriol
  Vinyals}, {and} \bibinfo{person}{Jeff Dean}.}
  \bibinfo{year}{2015}\natexlab{}.
\newblock \bibinfo{title}{Distilling the Knowledge in a Neural Network}.
\newblock
\newblock
\showeprint[arxiv]{1503.02531}~[stat.ML]


\bibitem[\protect\citeauthoryear{Hosseini, Yun, Park, Louizos, Soriaga, and
  Welling}{Hosseini et~al\mbox{.}}{2020}]%
        {TrainUA}
\bibfield{author}{\bibinfo{person}{Hossein Hosseini}, \bibinfo{person}{Sungrack
  Yun}, \bibinfo{person}{Hyunsin Park}, \bibinfo{person}{Christos Louizos},
  \bibinfo{person}{Joseph Soriaga}, {and} \bibinfo{person}{Max Welling}.}
  \bibinfo{year}{2020}\natexlab{}.
\newblock \bibinfo{title}{Federated Learning of User Authentication Models}.
\newblock
\newblock
\showeprint[arxiv]{2007.04618}~[cs.LG]


\bibitem[\protect\citeauthoryear{Huang and Liu}{Huang and Liu}{2019}]%
        {FederatedMedical}
\bibfield{author}{\bibinfo{person}{Li Huang} {and} \bibinfo{person}{Dianbo
  Liu}.} \bibinfo{year}{2019}\natexlab{}.
\newblock \showarticletitle{Patient Clustering Improves Efficiency of Federated
  Machine Learning to predict mortality and hospital stay time using
  distributed Electronic Medical Records}.
\newblock \bibinfo{journal}{\emph{CoRR}}  \bibinfo{volume}{abs/1903.09296}
  (\bibinfo{year}{2019}).
\newblock
\showeprint[arxiv]{1903.09296}
\urldef\tempurl%
\url{http://arxiv.org/abs/1903.09296}
\showURL{%
\tempurl}


\bibitem[\protect\citeauthoryear{Itahara, Nishio, Koda, Morikura, and
  Yamamoto}{Itahara et~al\mbox{.}}{2021}]%
        {ShareUnlabelledDataset}
\bibfield{author}{\bibinfo{person}{Sohei Itahara}, \bibinfo{person}{Takayuki
  Nishio}, \bibinfo{person}{Yusuke Koda}, \bibinfo{person}{Masahiro Morikura},
  {and} \bibinfo{person}{Koji Yamamoto}.} \bibinfo{year}{2021}\natexlab{}.
\newblock \bibinfo{title}{Distillation-Based Semi-Supervised Federated Learning
  for Communication-Efficient Collaborative Training with Non-IID Private
  Data}.
\newblock
\newblock
\showeprint[arxiv]{2008.06180}~[cs.DC]


\bibitem[\protect\citeauthoryear{Jeong, Yoon, Yang, and Hwang}{Jeong
  et~al\mbox{.}}{2020}]%
        {FedMatch}
\bibfield{author}{\bibinfo{person}{Wonyong Jeong}, \bibinfo{person}{Jaehong
  Yoon}, \bibinfo{person}{Eunho Yang}, {and} \bibinfo{person}{Sung~Ju Hwang}.}
  \bibinfo{year}{2020}\natexlab{}.
\newblock \bibinfo{title}{Federated Semi-Supervised Learning with Inter-Client
  Consistency}.
\newblock
\newblock
\showeprint[arxiv]{2006.12097}~[cs.LG]


\bibitem[\protect\citeauthoryear{Jin, Wei, Liu, and Yang}{Jin
  et~al\mbox{.}}{2020}]%
        {SSFL}
\bibfield{author}{\bibinfo{person}{Yilun Jin}, \bibinfo{person}{Xiguang Wei},
  \bibinfo{person}{Yang Liu}, {and} \bibinfo{person}{Qiang Yang}.}
  \bibinfo{year}{2020}\natexlab{}.
\newblock \bibinfo{title}{Towards Utilizing Unlabeled Data in Federated
  Learning: A Survey and Prospective}.
\newblock
\newblock
\showeprint[arxiv]{2002.11545}~[cs.LG]


\bibitem[\protect\citeauthoryear{Kairouz, McMahan, Avent, Bellet, Bennis,
  Bhagoji, Bonawitz, Charles, Cormode, Cummings, D'Oliveira, Eichner, Rouayheb,
  Evans, Gardner, Garrett, Gascón, Ghazi, Gibbons, Gruteser, Harchaoui, He,
  He, Huo, Hutchinson, Hsu, Jaggi, Javidi, Joshi, Khodak, Konečný, Korolova,
  Koushanfar, Koyejo, Lepoint, Liu, Mittal, Mohri, Nock, Özgür, Pagh,
  Raykova, Qi, Ramage, Raskar, Song, Song, Stich, Sun, Suresh, Tramèr,
  Vepakomma, Wang, Xiong, Xu, Yang, Yu, Yu, and Zhao}{Kairouz
  et~al\mbox{.}}{2021}]%
        {OpenChallenges}
\bibfield{author}{\bibinfo{person}{Peter Kairouz}, \bibinfo{person}{H.~Brendan
  McMahan}, \bibinfo{person}{Brendan Avent}, \bibinfo{person}{Aurélien
  Bellet}, \bibinfo{person}{Mehdi Bennis}, \bibinfo{person}{Arjun~Nitin
  Bhagoji}, \bibinfo{person}{Kallista Bonawitz}, \bibinfo{person}{Zachary
  Charles}, \bibinfo{person}{Graham Cormode}, \bibinfo{person}{Rachel
  Cummings}, \bibinfo{person}{Rafael G.~L. D'Oliveira}, \bibinfo{person}{Hubert
  Eichner}, \bibinfo{person}{Salim~El Rouayheb}, \bibinfo{person}{David Evans},
  \bibinfo{person}{Josh Gardner}, \bibinfo{person}{Zachary Garrett},
  \bibinfo{person}{Adrià Gascón}, \bibinfo{person}{Badih Ghazi},
  \bibinfo{person}{Phillip~B. Gibbons}, \bibinfo{person}{Marco Gruteser},
  \bibinfo{person}{Zaid Harchaoui}, \bibinfo{person}{Chaoyang He},
  \bibinfo{person}{Lie He}, \bibinfo{person}{Zhouyuan Huo},
  \bibinfo{person}{Ben Hutchinson}, \bibinfo{person}{Justin Hsu},
  \bibinfo{person}{Martin Jaggi}, \bibinfo{person}{Tara Javidi},
  \bibinfo{person}{Gauri Joshi}, \bibinfo{person}{Mikhail Khodak},
  \bibinfo{person}{Jakub Konečný}, \bibinfo{person}{Aleksandra Korolova},
  \bibinfo{person}{Farinaz Koushanfar}, \bibinfo{person}{Sanmi Koyejo},
  \bibinfo{person}{Tancrède Lepoint}, \bibinfo{person}{Yang Liu},
  \bibinfo{person}{Prateek Mittal}, \bibinfo{person}{Mehryar Mohri},
  \bibinfo{person}{Richard Nock}, \bibinfo{person}{Ayfer Özgür},
  \bibinfo{person}{Rasmus Pagh}, \bibinfo{person}{Mariana Raykova},
  \bibinfo{person}{Hang Qi}, \bibinfo{person}{Daniel Ramage},
  \bibinfo{person}{Ramesh Raskar}, \bibinfo{person}{Dawn Song},
  \bibinfo{person}{Weikang Song}, \bibinfo{person}{Sebastian~U. Stich},
  \bibinfo{person}{Ziteng Sun}, \bibinfo{person}{Ananda~Theertha Suresh},
  \bibinfo{person}{Florian Tramèr}, \bibinfo{person}{Praneeth Vepakomma},
  \bibinfo{person}{Jianyu Wang}, \bibinfo{person}{Li Xiong},
  \bibinfo{person}{Zheng Xu}, \bibinfo{person}{Qiang Yang},
  \bibinfo{person}{Felix~X. Yu}, \bibinfo{person}{Han Yu}, {and}
  \bibinfo{person}{Sen Zhao}.} \bibinfo{year}{2021}\natexlab{}.
\newblock \bibinfo{title}{Advances and Open Problems in Federated Learning}.
\newblock
\newblock
\showeprint[arxiv]{1912.04977}~[cs.LG]


\bibitem[\protect\citeauthoryear{Koizumi, Saito, Uematsu, Harada, and
  Imoto}{Koizumi et~al\mbox{.}}{2019}]%
        {MachineAnomalyDetection}
\bibfield{author}{\bibinfo{person}{Yuma Koizumi}, \bibinfo{person}{Shoichiro
  Saito}, \bibinfo{person}{Hisashi Uematsu}, \bibinfo{person}{Noboru Harada},
  {and} \bibinfo{person}{Keisuke Imoto}.} \bibinfo{year}{2019}\natexlab{}.
\newblock \bibinfo{title}{ToyADMOS: A Dataset of Miniature-Machine Operating
  Sounds for Anomalous Sound Detection}.
\newblock
\newblock
\showeprint[arxiv]{1908.03299}~[eess.AS]


\bibitem[\protect\citeauthoryear{Konečný, McMahan, Yu, Richtárik, Suresh,
  and Bacon}{Konečný et~al\mbox{.}}{2017}]%
        {FL}
\bibfield{author}{\bibinfo{person}{Jakub Konečný},
  \bibinfo{person}{H.~Brendan McMahan}, \bibinfo{person}{Felix~X. Yu},
  \bibinfo{person}{Peter Richtárik}, \bibinfo{person}{Ananda~Theertha Suresh},
  {and} \bibinfo{person}{Dave Bacon}.} \bibinfo{year}{2017}\natexlab{}.
\newblock \bibinfo{title}{Federated Learning: Strategies for Improving
  Communication Efficiency}.
\newblock
\newblock
\showeprint[arxiv]{1610.05492}~[cs.LG]


\bibitem[\protect\citeauthoryear{Korbar, Tran, and Torresani}{Korbar
  et~al\mbox{.}}{2018}]%
        {korbar2018cooperative}
\bibfield{author}{\bibinfo{person}{Bruno Korbar}, \bibinfo{person}{Du Tran},
  {and} \bibinfo{person}{Lorenzo Torresani}.} \bibinfo{year}{2018}\natexlab{}.
\newblock \showarticletitle{Cooperative learning of audio and video models from
  self-supervised synchronization}. In \bibinfo{booktitle}{\emph{Proceedings of
  the 32nd International Conference on Neural Information Processing Systems}}.
  \bibinfo{pages}{7774--7785}.
\newblock


\bibitem[\protect\citeauthoryear{Lake, Ullman, and andSamuel J.~Gershman}{Lake
  et~al\mbox{.}}{2016}]%
        {DeepLearningPerformance}
\bibfield{author}{\bibinfo{person}{Brenden~M. Lake}, \bibinfo{person}{Tomer~D.
  Ullman}, {and} \bibinfo{person}{Joshua B.~Tenenbaum andSamuel J.~Gershman}.}
  \bibinfo{year}{2016}\natexlab{}.
\newblock \showarticletitle{Building Machines That Learn and Think Like
  People}.
\newblock \bibinfo{journal}{\emph{CoRR}}  \bibinfo{volume}{abs/1604.00289}
  (\bibinfo{year}{2016}).
\newblock
\showeprint[arxiv]{1604.00289}
\urldef\tempurl%
\url{http://arxiv.org/abs/1604.00289}
\showURL{%
\tempurl}


\bibitem[\protect\citeauthoryear{Lee et~al\mbox{.}}{Lee et~al\mbox{.}}{2013}]%
        {PSL}
\bibfield{author}{\bibinfo{person}{Dong-Hyun Lee} {et~al\mbox{.}}}
  \bibinfo{year}{2013}\natexlab{}.
\newblock \showarticletitle{Pseudo-label: The simple and efficient
  semi-supervised learning method for deep neural networks}. In
  \bibinfo{booktitle}{\emph{Workshop on challenges in representation learning,
  ICML}}, Vol.~\bibinfo{volume}{3}.
\newblock


\bibitem[\protect\citeauthoryear{Leroy, Coucke, Lavril, Gisselbrecht, and
  Dureau}{Leroy et~al\mbox{.}}{2019}]%
        {KeywordSpottingIID}
\bibfield{author}{\bibinfo{person}{David Leroy}, \bibinfo{person}{Alice
  Coucke}, \bibinfo{person}{Thibaut Lavril}, \bibinfo{person}{Thibault
  Gisselbrecht}, {and} \bibinfo{person}{Joseph Dureau}.}
  \bibinfo{year}{2019}\natexlab{}.
\newblock \bibinfo{title}{Federated Learning for Keyword Spotting}.
\newblock
\newblock
\showeprint[arxiv]{1810.05512}~[eess.AS]


\bibitem[\protect\citeauthoryear{Li, Sahu, Talwalkar, and Smith}{Li
  et~al\mbox{.}}{2019}]%
        {FederatedSurvey}
\bibfield{author}{\bibinfo{person}{Tian Li}, \bibinfo{person}{Anit~Kumar Sahu},
  \bibinfo{person}{Ameet Talwalkar}, {and} \bibinfo{person}{Virginia Smith}.}
  \bibinfo{year}{2019}\natexlab{}.
\newblock \showarticletitle{Federated Learning: Challenges, Methods, and Future
  Directions}.
\newblock \bibinfo{journal}{\emph{CoRR}}  \bibinfo{volume}{abs/1908.07873}
  (\bibinfo{year}{2019}).
\newblock
\showeprint[arxiv]{1908.07873}
\urldef\tempurl%
\url{http://arxiv.org/abs/1908.07873}
\showURL{%
\tempurl}


\bibitem[\protect\citeauthoryear{Li, Sahu, Zaheer, Sanjabi, Talwalkar, and
  Smith}{Li et~al\mbox{.}}{2020}]%
        {FedProx}
\bibfield{author}{\bibinfo{person}{Tian Li}, \bibinfo{person}{Anit~Kumar Sahu},
  \bibinfo{person}{Manzil Zaheer}, \bibinfo{person}{Maziar Sanjabi},
  \bibinfo{person}{Ameet Talwalkar}, {and} \bibinfo{person}{Virginia Smith}.}
  \bibinfo{year}{2020}\natexlab{}.
\newblock \bibinfo{title}{Federated Optimization in Heterogeneous Networks}.
\newblock
\newblock
\showeprint[arxiv]{1812.06127}~[cs.LG]


\bibitem[\protect\citeauthoryear{Long, Che, Wang, Ye, Luo, Wu, Xiao, and
  Ma}{Long et~al\mbox{.}}{2020}]%
        {FedSemi}
\bibfield{author}{\bibinfo{person}{Zewei Long}, \bibinfo{person}{Liwei Che},
  \bibinfo{person}{Yaqing Wang}, \bibinfo{person}{Muchao Ye},
  \bibinfo{person}{Junyu Luo}, \bibinfo{person}{Jinze Wu},
  \bibinfo{person}{Houping Xiao}, {and} \bibinfo{person}{Fenglong Ma}.}
  \bibinfo{year}{2020}\natexlab{}.
\newblock \bibinfo{title}{FedSemi: An Adaptive Federated Semi-Supervised
  Learning Framework}.
\newblock
\newblock
\showeprint[arxiv]{2012.03292}~[cs.LG]


\bibitem[\protect\citeauthoryear{Loshchilov and Hutter}{Loshchilov and
  Hutter}{2016}]%
        {cosine_sceduler}
\bibfield{author}{\bibinfo{person}{Ilya Loshchilov} {and}
  \bibinfo{person}{Frank Hutter}.} \bibinfo{year}{2016}\natexlab{}.
\newblock \showarticletitle{{SGDR:} Stochastic Gradient Descent with Restarts}.
\newblock \bibinfo{journal}{\emph{CoRR}}  \bibinfo{volume}{abs/1608.03983}
  (\bibinfo{year}{2016}).
\newblock
\showeprint[arXiv]{1608.03983}
\urldef\tempurl%
\url{http://arxiv.org/abs/1608.03983}
\showURL{%
\tempurl}


\bibitem[\protect\citeauthoryear{Mac~Aodha, Gibb, Barlow, Browning, Firman,
  Freeman, Harder, Kinsey, Mead, Newson, Pandourski, Parsons, Russ,
  Szodoray-Paradi, Szodoray-Paradi, Tilova, Girolami, Brostow, and
  Jones}{Mac~Aodha et~al\mbox{.}}{2018}]%
        {BatDetection}
\bibfield{author}{\bibinfo{person}{Oisin Mac~Aodha}, \bibinfo{person}{Rory
  Gibb}, \bibinfo{person}{Kate~E. Barlow}, \bibinfo{person}{Ella Browning},
  \bibinfo{person}{Michael Firman}, \bibinfo{person}{Robin Freeman},
  \bibinfo{person}{Briana Harder}, \bibinfo{person}{Libby Kinsey},
  \bibinfo{person}{Gary~R. Mead}, \bibinfo{person}{Stuart~E. Newson},
  \bibinfo{person}{Ivan Pandourski}, \bibinfo{person}{Stuart Parsons},
  \bibinfo{person}{Jon Russ}, \bibinfo{person}{Abigel Szodoray-Paradi},
  \bibinfo{person}{Farkas Szodoray-Paradi}, \bibinfo{person}{Elena Tilova},
  \bibinfo{person}{Mark Girolami}, \bibinfo{person}{Gabriel Brostow}, {and}
  \bibinfo{person}{Kate~E. Jones}.} \bibinfo{year}{2018}\natexlab{}.
\newblock \showarticletitle{Bat detective—Deep learning tools for bat
  acoustic signal detection}.
\newblock \bibinfo{journal}{\emph{PLOS Computational Biology}}
  \bibinfo{volume}{14}, \bibinfo{number}{3} (\bibinfo{date}{03}
  \bibinfo{year}{2018}), \bibinfo{pages}{1--19}.
\newblock
\urldef\tempurl%
\url{https://doi.org/10.1371/journal.pcbi.1005995}
\showDOI{\tempurl}


\bibitem[\protect\citeauthoryear{MacLean}{MacLean}{2018}]%
        {VF}
\bibfield{author}{\bibinfo{person}{Ken MacLean}.}
  \bibinfo{year}{2018}\natexlab{}.
\newblock \showarticletitle{Voxforge}.
\newblock \bibinfo{journal}{\emph{Ken MacLean.[Online]. Available:
  http://www.voxforge.org/home.[Acedido em 2012]}} (\bibinfo{year}{2018}).
\newblock


\bibitem[\protect\citeauthoryear{McMahan, Moore, Ramage, Hampson, and
  y~Arcas}{McMahan et~al\mbox{.}}{2017}]%
        {FedAvg}
\bibfield{author}{\bibinfo{person}{H.~Brendan McMahan}, \bibinfo{person}{Eider
  Moore}, \bibinfo{person}{Daniel Ramage}, \bibinfo{person}{Seth Hampson},
  {and} \bibinfo{person}{Blaise~Agüera y Arcas}.}
  \bibinfo{year}{2017}\natexlab{}.
\newblock \bibinfo{title}{Communication-Efficient Learning of Deep Networks
  from Decentralized Data}.
\newblock
\newblock
\showeprint[arxiv]{1602.05629}~[cs.LG]


\bibitem[\protect\citeauthoryear{Miyato, ichi Maeda, Koyama, and Ishii}{Miyato
  et~al\mbox{.}}{2018}]%
        {VAT}
\bibfield{author}{\bibinfo{person}{Takeru Miyato}, \bibinfo{person}{Shin ichi
  Maeda}, \bibinfo{person}{Masanori Koyama}, {and} \bibinfo{person}{Shin
  Ishii}.} \bibinfo{year}{2018}\natexlab{}.
\newblock \bibinfo{title}{Virtual Adversarial Training: A Regularization Method
  for Supervised and Semi-Supervised Learning}.
\newblock
\newblock
\showeprint[arxiv]{1704.03976}~[stat.ML]


\bibitem[\protect\citeauthoryear{Oliver, Odena, Raffel, Cubuk, and
  Goodfellow}{Oliver et~al\mbox{.}}{2019}]%
        {SSLRealisticEvaluation}
\bibfield{author}{\bibinfo{person}{Avital Oliver}, \bibinfo{person}{Augustus
  Odena}, \bibinfo{person}{Colin Raffel}, \bibinfo{person}{Ekin~D. Cubuk},
  {and} \bibinfo{person}{Ian~J. Goodfellow}.} \bibinfo{year}{2019}\natexlab{}.
\newblock \bibinfo{title}{Realistic Evaluation of Deep Semi-Supervised Learning
  Algorithms}.
\newblock
\newblock
\showeprint[arxiv]{1804.09170}~[cs.LG]


\bibitem[\protect\citeauthoryear{Park, Min, Bhattacharya, and Kawsar}{Park
  et~al\mbox{.}}{2020}]%
        {ABCX}
\bibfield{author}{\bibinfo{person}{Chunjong Park}, \bibinfo{person}{Chulhong
  Min}, \bibinfo{person}{Sourav Bhattacharya}, {and} \bibinfo{person}{Fahim
  Kawsar}.} \bibinfo{year}{2020}\natexlab{}.
\newblock \showarticletitle{Augmenting Conversational Agents with Ambient
  Acoustic Contexts}. In \bibinfo{booktitle}{\emph{22nd International
  Conference on Human-Computer Interaction with Mobile Devices and Services}}
  (Oldenburg, Germany) \emph{(\bibinfo{series}{MobileHCI '20})}.
  \bibinfo{publisher}{Association for Computing Machinery},
  \bibinfo{address}{New York, NY, USA}, Article \bibinfo{articleno}{33},
  \bibinfo{numpages}{9}~pages.
\newblock
\showISBNx{9781450375160}
\urldef\tempurl%
\url{https://doi.org/10.1145/3379503.3403535}
\showDOI{\tempurl}


\bibitem[\protect\citeauthoryear{Ramaswamy, Mathews, Rao, and
  Beaufays}{Ramaswamy et~al\mbox{.}}{2019}]%
        {FederatedEmojiPrediction}
\bibfield{author}{\bibinfo{person}{Swaroop Ramaswamy}, \bibinfo{person}{Rajiv
  Mathews}, \bibinfo{person}{Kanishka Rao}, {and}
  \bibinfo{person}{Fran{\c{c}}oise Beaufays}.} \bibinfo{year}{2019}\natexlab{}.
\newblock \showarticletitle{Federated Learning for Emoji Prediction in a Mobile
  Keyboard}.
\newblock \bibinfo{journal}{\emph{CoRR}}  \bibinfo{volume}{abs/1906.04329}
  (\bibinfo{year}{2019}).
\newblock
\showeprint[arxiv]{1906.04329}
\urldef\tempurl%
\url{http://arxiv.org/abs/1906.04329}
\showURL{%
\tempurl}


\bibitem[\protect\citeauthoryear{Saeed, Grangier, and Zeghidour}{Saeed
  et~al\mbox{.}}{2021}]%
        {saeed2021contrastive}
\bibfield{author}{\bibinfo{person}{Aaqib Saeed}, \bibinfo{person}{David
  Grangier}, {and} \bibinfo{person}{Neil Zeghidour}.}
  \bibinfo{year}{2021}\natexlab{}.
\newblock \showarticletitle{Contrastive learning of general-purpose audio
  representations}. In \bibinfo{booktitle}{\emph{ICASSP 2021-2021 IEEE
  International Conference on Acoustics, Speech and Signal Processing
  (ICASSP)}}. IEEE, \bibinfo{pages}{3875--3879}.
\newblock


\bibitem[\protect\citeauthoryear{Stowell, Stylianou, Wood, Pamula, and
  Glotin}{Stowell et~al\mbox{.}}{2018}]%
        {BirdDetection}
\bibfield{author}{\bibinfo{person}{Dan Stowell}, \bibinfo{person}{Yannis
  Stylianou}, \bibinfo{person}{Mike Wood}, \bibinfo{person}{Hanna Pamula},
  {and} \bibinfo{person}{Herv{\'{e}} Glotin}.} \bibinfo{year}{2018}\natexlab{}.
\newblock \showarticletitle{Automatic acoustic detection of birds through deep
  learning: the first Bird Audio Detection challenge}.
\newblock \bibinfo{journal}{\emph{CoRR}}  \bibinfo{volume}{abs/1807.05812}
  (\bibinfo{year}{2018}).
\newblock
\showeprint[arxiv]{1807.05812}
\urldef\tempurl%
\url{http://arxiv.org/abs/1807.05812}
\showURL{%
\tempurl}


\bibitem[\protect\citeauthoryear{Tagliasacchi, Gfeller, Quitry, and
  Roblek}{Tagliasacchi et~al\mbox{.}}{2019}]%
        {Model}
\bibfield{author}{\bibinfo{person}{Marco Tagliasacchi}, \bibinfo{person}{Beat
  Gfeller}, \bibinfo{person}{F{\'e}lix de~Chaumont Quitry}, {and}
  \bibinfo{person}{Dominik Roblek}.} \bibinfo{year}{2019}\natexlab{}.
\newblock \showarticletitle{Self-supervised audio representation learning for
  mobile devices}.
\newblock \bibinfo{journal}{\emph{arXiv preprint arXiv:1905.11796}}
  (\bibinfo{year}{2019}).
\newblock


\bibitem[\protect\citeauthoryear{Tarvainen and Valpola}{Tarvainen and
  Valpola}{2018}]%
        {MT}
\bibfield{author}{\bibinfo{person}{Antti Tarvainen} {and}
  \bibinfo{person}{Harri Valpola}.} \bibinfo{year}{2018}\natexlab{}.
\newblock \bibinfo{title}{Mean teachers are better role models: Weight-averaged
  consistency targets improve semi-supervised deep learning results}.
\newblock
\newblock
\showeprint[arxiv]{1703.01780}~[cs.NE]


\bibitem[\protect\citeauthoryear{van Berlo, Saeed, and Ozcelebi}{van Berlo
  et~al\mbox{.}}{2020}]%
        {ActivityRecognition}
\bibfield{author}{\bibinfo{person}{Bram van Berlo}, \bibinfo{person}{Aaqib
  Saeed}, {and} \bibinfo{person}{Tanir Ozcelebi}.}
  \bibinfo{year}{2020}\natexlab{}.
\newblock \bibinfo{title}{Towards federated unsupervised representation
  learning}.
\newblock , \bibinfo{numpages}{31--36}~pages.
\newblock


\bibitem[\protect\citeauthoryear{van Engelen and Hoos}{van Engelen and
  Hoos}{2019}]%
        {SSLSurvey}
\bibfield{author}{\bibinfo{person}{Jesper~E. van Engelen} {and}
  \bibinfo{person}{H. Hoos}.} \bibinfo{year}{2019}\natexlab{}.
\newblock \showarticletitle{A survey on semi-supervised learning}.
\newblock \bibinfo{journal}{\emph{Machine Learning}}  \bibinfo{volume}{109}
  (\bibinfo{year}{2019}), \bibinfo{pages}{373--440}.
\newblock


\bibitem[\protect\citeauthoryear{Warden}{Warden}{2018}]%
        {SPCM}
\bibfield{author}{\bibinfo{person}{Pete Warden}.}
  \bibinfo{year}{2018}\natexlab{}.
\newblock \showarticletitle{Speech Commands: {A} Dataset for Limited-Vocabulary
  Speech Recognition}.
\newblock \bibinfo{journal}{\emph{CoRR}}  \bibinfo{volume}{abs/1804.03209}
  (\bibinfo{year}{2018}).
\newblock
\showeprint[arxiv]{1804.03209}
\urldef\tempurl%
\url{http://arxiv.org/abs/1804.03209}
\showURL{%
\tempurl}


\bibitem[\protect\citeauthoryear{Wu and He}{Wu and He}{2018}]%
        {GNorm}
\bibfield{author}{\bibinfo{person}{Yuxin Wu} {and} \bibinfo{person}{Kaiming
  He}.} \bibinfo{year}{2018}\natexlab{}.
\newblock \bibinfo{title}{Group Normalization}.
\newblock
\newblock
\showeprint[arxiv]{1803.08494}~[cs.CV]


\bibitem[\protect\citeauthoryear{Xie, Hovy, Luong, and Le}{Xie
  et~al\mbox{.}}{2019}]%
        {teacher_Student}
\bibfield{author}{\bibinfo{person}{Qizhe Xie}, \bibinfo{person}{Eduard~H.
  Hovy}, \bibinfo{person}{Minh{-}Thang Luong}, {and} \bibinfo{person}{Quoc~V.
  Le}.} \bibinfo{year}{2019}\natexlab{}.
\newblock \showarticletitle{Self-training with Noisy Student improves ImageNet
  classification}.
\newblock \bibinfo{journal}{\emph{CoRR}}  \bibinfo{volume}{abs/1911.04252}
  (\bibinfo{year}{2019}).
\newblock
\showeprint[arXiv]{1911.04252}
\urldef\tempurl%
\url{http://arxiv.org/abs/1911.04252}
\showURL{%
\tempurl}


\bibitem[\protect\citeauthoryear{Yang, Andrew, Eichner, Sun, Li, Kong, Ramage,
  and Beaufays}{Yang et~al\mbox{.}}{2018}]%
        {KeywordSuggestion}
\bibfield{author}{\bibinfo{person}{Timothy Yang}, \bibinfo{person}{Galen
  Andrew}, \bibinfo{person}{Hubert Eichner}, \bibinfo{person}{Haicheng Sun},
  \bibinfo{person}{Wei Li}, \bibinfo{person}{Nicholas Kong},
  \bibinfo{person}{Daniel Ramage}, {and} \bibinfo{person}{Françoise
  Beaufays}.} \bibinfo{year}{2018}\natexlab{}.
\newblock \bibinfo{title}{Applied Federated Learning: Improving Google Keyboard
  Query Suggestions}.
\newblock
\newblock
\showeprint[arxiv]{1812.02903}~[cs.LG]


\bibitem[\protect\citeauthoryear{Zhao, Li, Lai, Suda, Civin, and Chandra}{Zhao
  et~al\mbox{.}}{2018}]%
        {NonIID}
\bibfield{author}{\bibinfo{person}{Yue Zhao}, \bibinfo{person}{Meng Li},
  \bibinfo{person}{Liangzhen Lai}, \bibinfo{person}{Naveen Suda},
  \bibinfo{person}{Damon Civin}, {and} \bibinfo{person}{Vikas Chandra}.}
  \bibinfo{year}{2018}\natexlab{}.
\newblock \bibinfo{title}{Federated Learning with Non-IID Data}.
\newblock
\newblock
\showeprint[arxiv]{1806.00582}~[cs.LG]


\bibitem[\protect\citeauthoryear{Zhu and Goldberg}{Zhu and Goldberg}{2009}]%
        {SSL}
\bibfield{author}{\bibinfo{person}{Xiaojin Zhu} {and} \bibinfo{person}{Andrew
  Goldberg}.} \bibinfo{year}{2009}\natexlab{}.
\newblock \bibinfo{booktitle}{\emph{Introduction to Semi-Supervised Learning}}.
  Vol.~\bibinfo{volume}{3}.
\newblock
\urldef\tempurl%
\url{https://doi.org/10.2200/S00196ED1V01Y200906AIM006}
\showDOI{\tempurl}


\end{thebibliography}
\newpage
\appendix
\section*{APPENDIX}
\begin{table*}[!htbp]
\centering \small
\caption{Performance evaluation of \method~ when varying both labeled and unlabeled datasets. Average accuracy over 3 distinct runs is reported on Speech Commands, including variance across experiments. Federated parameters are set to $q$=80\%, $\sigma$=25\%, $\beta$=0.5, $R$=100, $E$=1, $N$=15.} \label{tab:vary_u_std}
\resizebox{\textwidth}{!}{%
    \begin{tabular}{l cccc cccc}
        \toprule
        \multicolumn{1}{c}{\multirow{2}{*}{\begin{tabular}[c]{@{}l@{}}\textbf{Labeled}\\\textbf{Percentage}\end{tabular}}} & \multicolumn{4}{c}{\textbf{FedSTAR (Randomly Initialized)}} & \multicolumn{4}{c}{\textbf{FedSTAR (Pre-Trained)}} \\
        \cmidrule[0.5pt](rl){2-5} \cmidrule[0.5pt](rl){6-9}
        & $U$=20\% & $U$=50\% & $U$=80\% & $U$=100\% & $U$=20\% & $U$=50\% & $U$=80\% & $U$=100\% \\
        \midrule[0.5pt]
        $L$=3\%   & 84.13 $\pm$ 0.004 & 85.40 $\pm$ 0.008 & 86.63 $\pm$ 0.002 & 86.82 $\pm$ 0.020 
                  & 84.52 $\pm$ 0.001 & 85.17 $\pm$ 0.001 & 85.43 $\pm$ 0.001 & 86.46 $\pm$ 0.006 \\
        $L$=5\%   & 87.47 $\pm$ 0.001 & 88.52 $\pm$ 0.005 & 88.90 $\pm$ 0.001 & 89.33 $\pm$ 0.007	
                  & 88.07 $\pm$ 0.004 & 88.28 $\pm$ 0.002 & 87.73 $\pm$ 0.001 & 89.98 $\pm$ 0.002 \\  
        $L$=20\%  & 90.06 $\pm$ 0.003 & 92.24 $\pm$ 0.012 & 93.07 $\pm$ 0.001 & 93.15 $\pm$ 0.011	
                  & 92.44 $\pm$ 0.001 & 93.67 $\pm$ 0.005 & 93.98 $\pm$ 0.003 & 94.13 $\pm$ 0.001 \\
        $L$=50\%  & 87.76 $\pm$ 0.003 & 92.26 $\pm$ 0.005 & 94.18 $\pm$ 0.001 & 93.38 $\pm$ 0.001	
                  & 90.70 $\pm$ 0.001 & 93.83 $\pm$ 0.003 & 94.76 $\pm$ 0.001 & 95.54 $\pm$ 0.007 \\
        \bottomrule
    \end{tabular}%
}
\end{table*}

\begin{table*}[!htbp]
\centering \small
\caption{Performance evaluation of method~ against variation of class availability across clients. Class distribution has mean $\mu$=3 and variance $\sigma_{c}$. Average accuracy over 3 distinct runs is reported on Speech Commands, including variance across experiments. Federated parameters are set to $\beta=0.5$, $R$=100, $N$=15, $q$=80\% and $E$=1.} \label{tab:c_dist_std}
\resizebox{\textwidth}{!}{%
\begin{tabular}{lc cccccccc}
    \toprule
    \multicolumn{2}{c}{\multirow{2}{*}{\begin{tabular}[c]{@{}c@{}}\textbf{Class Distribution}\\ \textbf{Characteristics}\end{tabular}}}
    & \multicolumn{4}{c}{\textbf{Supervised (Federated)}} 
    & \multicolumn{4}{c}{\textbf{FedSTAR}} \\ 
    \cmidrule[0.5pt](rl){3-6} \cmidrule[0.5pt](rl){7-10} 
    &   & $L=3\%$ & $L=5\%$ & $L=20\%$ & $L=50\%$
        & $L=3\%$ & $L=5\%$ & $L=20\%$ & $L=50\%$ \\ 

    \midrule[0.5pt]
    \multirow{3}{*}{$\mu$=3} & $\sigma_{c}$=0 \%
                & 9.83  $\pm$ 0.017 & 32.63 $\pm$ 0.097 & 80.22 $\pm$ 0.056 & 82.40 0.048
                & 79.08 $\pm$ 0.026 & 79.62 $\pm$ 0.034 & 87.01 $\pm$ 0.028 & 83.14 $\pm$ 0.069 \\

    & $\sigma_{c}$=25\%
                & 10.54 $\pm$ 0.016 & 23.97 $\pm$ 0.139 & 75.41 $\pm$ 0.055 & 83.61 $\pm$ 0.046
                & 79.05 $\pm$ 0.052 & 84.15 $\pm$ 0.013 & 86.52 $\pm$ 0.032 & 85.05 $\pm$ 0.051 \\

    & $\sigma_{c}$=50\%     
                & 8.44  $\pm$ 0.001 & 24.25 $\pm$ 0.140 & 73.93 $\pm$ 0.044 & 84.41 $\pm$ 0.043 
                & 78.14 $\pm$ 0.021 & 81.88 $\pm$ 0.031 & 84.56 $\pm$ 0.041 & 84.55 $\pm$ 0.055 \\
    \bottomrule
\end{tabular}%
}
\end{table*}

\begin{landscape}
\begin{table*}[h]
\centering 
\small
\caption{Performance evaluation of \method. Average accuracy over 3 distinct trials on test set is reported, including variance across experiments. Federated parameters are set to $q$=80\%, $\sigma$=25\%, $\beta$=0.5, $E$=1, $R$=100.} \label{tab:results_std}
\begin{tabular}{l p{1cm} cccccccc}
    \toprule

    \multicolumn{1}{c}{\multirow{2}{*}{\textbf{Dataset}}} 
    & \multicolumn{1}{c}{\multirow{2}{*}{\textbf{Clients}}}
    & \multicolumn{4}{c}{\textbf{Supervised (Federated)}} 
    & \multicolumn{4}{c}{\textbf{FedSTAR}} \\ 
    \cmidrule[0.5pt](rl){3-6} \cmidrule[0.5pt](rl){7-10}

    & & $L=3\%$ & $L=5\%$ & $L=20\%$ & $L=50\%$
    & $L=3\%$ & $L=5\%$ & $L=20\%$ & $L=50\%$ \\ 

    \midrule[0.5pt]
    Ambient Context & \multicolumn{1}{c}{\multirow{3}{*}{$5$}}
                        & 46.34 $\pm$ 0.009 & 47.89 $\pm$ 0.056 & 61.40 $\pm$ 0.001 & 65.85 $\pm$ 0.021
                        & 48.68 $\pm$ 0.004 & 54.95 $\pm$ 0.026 & 64.37 $\pm$ 0.012 & 67.04 $\pm$ 0.010 \\

    Speech Commands &   & 81.12 $\pm$ 0.037 & 87.97  $\pm$ 0.047 & 92.35 $\pm$ 0.030 & 94.66 $\pm$ 0.012
                        & 87.41 $\pm$ 0.007 & 90.01  $\pm$ 0.001 & 94.17 $\pm$ 0.003 & 94.85 $\pm$ 0.001 \\

    VoxForge        &   & 54.55 $\pm$ 0.009 & 56.41 $\pm$ 0.021 & 61.65 $\pm$ 0.005 & 70.37 $\pm$ 0.021
                        & 63.92 $\pm$ 0.016 & 67.80 $\pm$ 0.018 & 69.09 $\pm$ 0.013 & 67.08 $\pm$ 0.016 \\
    \midrule[0.5pt]

    Ambient Context & \multicolumn{1}{c}{\multirow{3}{*}{$10$}}
                        & 35.29 $\pm$ 0.006 & 41.31 $\pm$ 0.012 & 51.71 $\pm$ 0.009 & 62.69 $\pm$ 0.018
                        & 48.87 $\pm$ 0.004 & 52.37 $\pm$ 0.018 & 62.94 $\pm$ 0.024 & 64.42 $\pm$ 0.006 \\
                        
    Speech Commands &   & 67.75 $\pm$ 0.001 & 83.80 $\pm$ 0.029 & 92.12 $\pm$ 0.087 & 94.02 $\pm$ 0.036
                        & 86.82 $\pm$ 0.006 & 90.33 $\pm$ 0.007 & 94.09 $\pm$ 0.002 & 94.18 $\pm$ 0.006 \\
                        
    VoxForge        &   & 56.14 $\pm$ 0.020 & 54.73 $\pm$ 0.001 & 60.48 $\pm$ 0.033 & 62.41 $\pm$ 0.014
                        & 59.87 $\pm$ 0.024 & 64.35 $\pm$ 0.003 & 69.38 $\pm$ 0.016 & 63.27 $\pm$ 0.032 \\
    \midrule[0.5pt]
    
    Ambient Context & \multicolumn{1}{c}{\multirow{3}{*}{$15$}}
                        & 33.03 $\pm$ 0.002 & 42.75 $\pm$ 0.007 & 53.37 $\pm$ 0.004 & 59.97 $\pm$ 0.004
                        & 49.54 $\pm$ 0.005 & 54.71 $\pm$ 0.022 & 63.46 $\pm$ 0.004 & 62.41 $\pm$ 0.006 \\
                        
    Speech Commands &   & 62.98 $\pm$ 0.003 & 72.84 $\pm$ 0.001 & 92.14 $\pm$ 0.003 & 93.14 $\pm$ 0.004
                        & 86.82 $\pm$ 0.006 & 89.33 $\pm$ 0.002 & 93.16 $\pm$ 0.001 & 93.39 $\pm$ 0.007 \\
                        
    VoxForge        &   & 54.26 $\pm$ 0.002 & 54.37 $\pm$ 0.009 & 57.11 $\pm$ 0.031 & 60.29 $\pm$ 0.001
                        & 55.82 $\pm$ 0.011 & 57.96 $\pm$ 0.025 & 67.66 $\pm$ 0.004 & 61.66 $\pm$ 0.007 \\
    \midrule[0.5pt]

    Ambient Context & \multicolumn{1}{c}{\multirow{3}{*}{$30$}}
                        & 32.31 $\pm$ 0.004 & 40.17 $\pm$ 0.001 & 47.05 $\pm$ 0.001 & 55.85 $\pm$ 0.002
                        & 40.84 $\pm$ 0.041 & 46.58 $\pm$ 0.013 & 60.21 $\pm$ 0.013 & 56.19 $\pm$ 0.009 \\
                        
    Speech Commands &   & 33.78 $\pm$ 0.012 & 44.21 $\pm$ 0.016 & 84.94 $\pm$ 0.012 & 92.21 $\pm$ 0.008
                        & 83.88 $\pm$ 0.001 & 88.19 $\pm$ 0.005 & 92.92 $\pm$ 0.005 & 92.62 $\pm$ 0.007 \\
                        
    VoxForge        &   & 50.32 $\pm$ 0.009 & 54.33 $\pm$ 0.015 & 55.19 $\pm$ 0.011 & 57.56 $\pm$ 0.002
                        & 54.81 $\pm$ 0.001 & 56.18 $\pm$ 0.005 & 63.83 $\pm$ 0.009 & 56.66 $\pm$ 0.009 \\
    \bottomrule

\end{tabular}%
\end{table*}
\end{landscape}

\end{document}